%% file: main.tex
\documentclass[10pt,twocolumn,letterpaper]{article}

\usepackage[pagenumbers]{cvpr} 


\definecolor{cvprblue}{rgb}{0.21,0.49,0.74}
\definecolor{Light}{rgb}{0.99, 0.92, 0.95}
\definecolor{codered}{rgb}{0.98,0.49,0.72}
\usepackage[pagebackref,breaklinks,colorlinks,allcolors=cvprblue]{hyperref}
\usepackage{amsmath}
\usepackage{graphicx}
\usepackage{amsfonts}
\usepackage{multirow}
\usepackage{array}
\usepackage{makecell}
\usepackage{epsfig}
\usepackage{tocloft}
\usepackage{colortbl}

\def\paperID{3246} 
\def\confName{CVPR}
\def\confYear{2025}
\definecolor{c1}{HTML}{70f3ff} 

\title{CoE: Chain-of-Explanation via Automatic Visual Concept Circuit Description and Polysemanticity Quantification}

\author{%
Wenlong Yu~~~~Qilong Wang\thanks{Corresponding author. Published in CVPR 2025.}~~~~Chuang Liu~~~~Dong Li~~~~Qinghua Hu \\
Tianjin Key Lab of Machine Learning, College of Intelligence and Computing, Tianjin University\\
 \small{\texttt{\{ywl\_95, qlwang, liuc\_09, li\_dong, huqinghua\}@tju.edu.cn}}\\
}

\begin{document}
\maketitle

\begin{abstract}
Explainability is a critical factor influencing the wide deployment of deep vision models (DVMs). 
Concept-based post-hoc explanation methods can provide both global and local insights into model decisions. 
However, current methods in this field face challenges in that they are inflexible to automatically construct accurate and sufficient linguistic explanations for global concepts and local circuits. 
Particularly, the intrinsic polysemanticity in semantic Visual Concepts (VCs) impedes the interpretability of concepts and DVMs, which is underestimated severely.
In this paper, we propose a \textbf{C}hain-\textbf{o}f-\textbf{E}xplanation (\textbf{CoE}) approach to address these issues.
Specifically, CoE automates the decoding and description of VCs 
to construct global concept explanation datasets. 
Further, to alleviate the effect of polysemanticity on model explainability, we design a concept polysemanticity disentanglement and filtering mechanism to distinguish the most contextually relevant concept atoms. 
Besides, a Concept Polysemanticity Entropy (CPE), as a measure of model interpretability, is formulated to quantify the degree of concept uncertainty.
The modeling of deterministic concepts is upgraded to uncertain concept atom distributions. 
Finally, CoE automatically enables linguistic local explanations of the decision-making process of DVMs by tracing the concept circuit. 
GPT-4o and human-based experiments demonstrate the effectiveness of CPE and the superiority of CoE, achieving an average absolute improvement of 36\% in terms of explainability scores.
\end{abstract}

\begin{figure}
\centering
    \includegraphics[width=0.95\columnwidth]{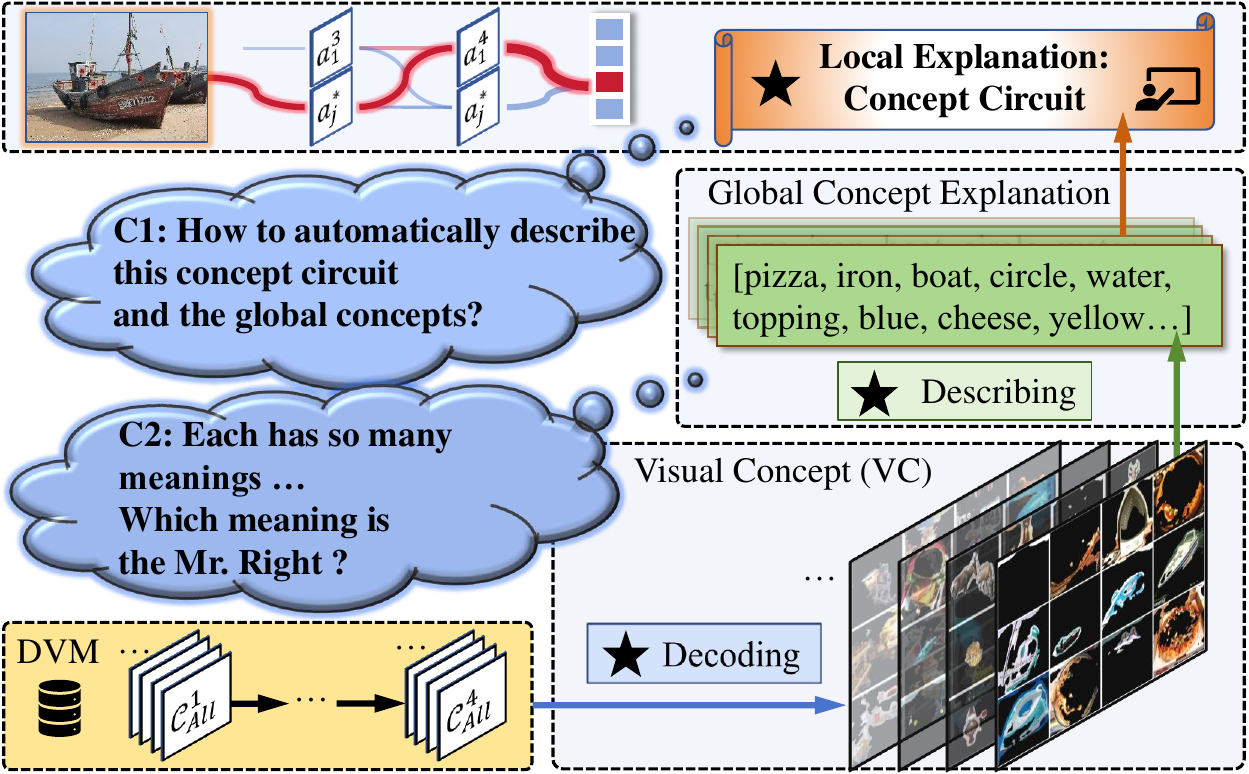}
\caption{Illustration of the integrated process of concept-based post-hoc XAI methods \cite{lrpsurvey}, along with two challenges they face.
}
\vskip -0.15in
\label{fig.egconcept}
\end{figure}

\addtocontents{toc}{\protect\setcounter{tocdepth}{0}}

\section{Introduction}
\label{sec.intro}

Deep learning-based vision models have demonstrated outstanding performance in various visual tasks, owing to the powerful learning and representation capabilities of deep neural networks (DNNs) \cite{lecun2015deep, pami-disen-suevey}. 
However, the extreme complexity of DNNs, while enhancing their performance, also limits their explainability \cite{xaisurveypami}. 
Humans are often unaware of the reasons behind the model's conclusions. This lack of explainability significantly hinders the wide application of deep vision models (DVMs) in critical fields such as autonomous driving and medical diagnosis \cite{xaisurveyauto, xaisurveymedicaltai}.

Post-hoc eXplainable Artificial Intelligence (XAI) can enhance the transparency of DVMs without causing degradations in performance \cite{xaisurveykbs, xaisurveyair}.
Research in this field can roughly be divided into local and global XAI methods \cite{xaisurvey1}.
The former explains either specific stages or the entire decision path with respect to a specific sample, identifying key activated regions or significant semantic concepts \cite{explainanything, gradcam, visioncircuit}.
The latter deciphers the overall behavior of the DVM in a dataset by concept discovery or visualization techniques \cite{tcav, autoconcept, globalxai}.
Among them, concept-based methods can explain DVMs from both global and local (glocal) perspectives \cite{lrp,naturecrp}, assuming that each channel or neuron in a certain layer is responsible for recognizing essential semantic concepts.
As shown in Fig. \ref{fig.egconcept}, they primarily perform one or two of the three operations: decoding Visual Concepts (VCs) of DVMs, manually constructing global concept explanation datasets, and providing local concept circuit explanations for individual samples \cite{milan, devil, pure}.

Despite the significant progress made in concept-based explanation research, certain challenges remain, as illustrated in Fig. \ref{fig.egconcept}.
First, the VCs are decoded by manually identifying and annotating the commonalities of a set of extracted image patches \cite{naturecrp, milan}.
These manual methods intuitively become ineffective when faced with new scenarios, new models, and new XAI methods, due to high labor costs, low efficiency, and the limited cognitive scope of individual annotators.
Second, most methods make a naive assumption: each concept is monosemantic and user-friendly \cite{understandingpoly}.
However, in practice, VCs encapsulate polysemantic information \cite{nature_fear1} (\eg, an average of 13 concept atoms are coupled per channel in a ResNet152 model), presenting challenges for the model explanation.
Particularly, polysemanticity can serve as a metric for evaluating a model's interpretability.
A higher level of polysemanticity signifies a broader conceptual space, increasing the risks of selecting incorrect semantics and generating deviated explanations.
Arbitrary concept selection or presenting them solely as images without descriptions yields suboptimal explanations \cite{xaisurveykbs}. 
Both visualizations and linguistic explanations become challenging for humans to comprehend, as it is unclear which semantic is actually represented or holds prominence \cite{pure}. 
Although polysemanticity is crucial to XAI, its systematic quantification and resolution remain unaddressed.
All of these challenges collectively exacerbate the difficulty of concept-based explanations for DVMs.

\begin{figure*}
\centering
    \includegraphics[width=0.99\linewidth]{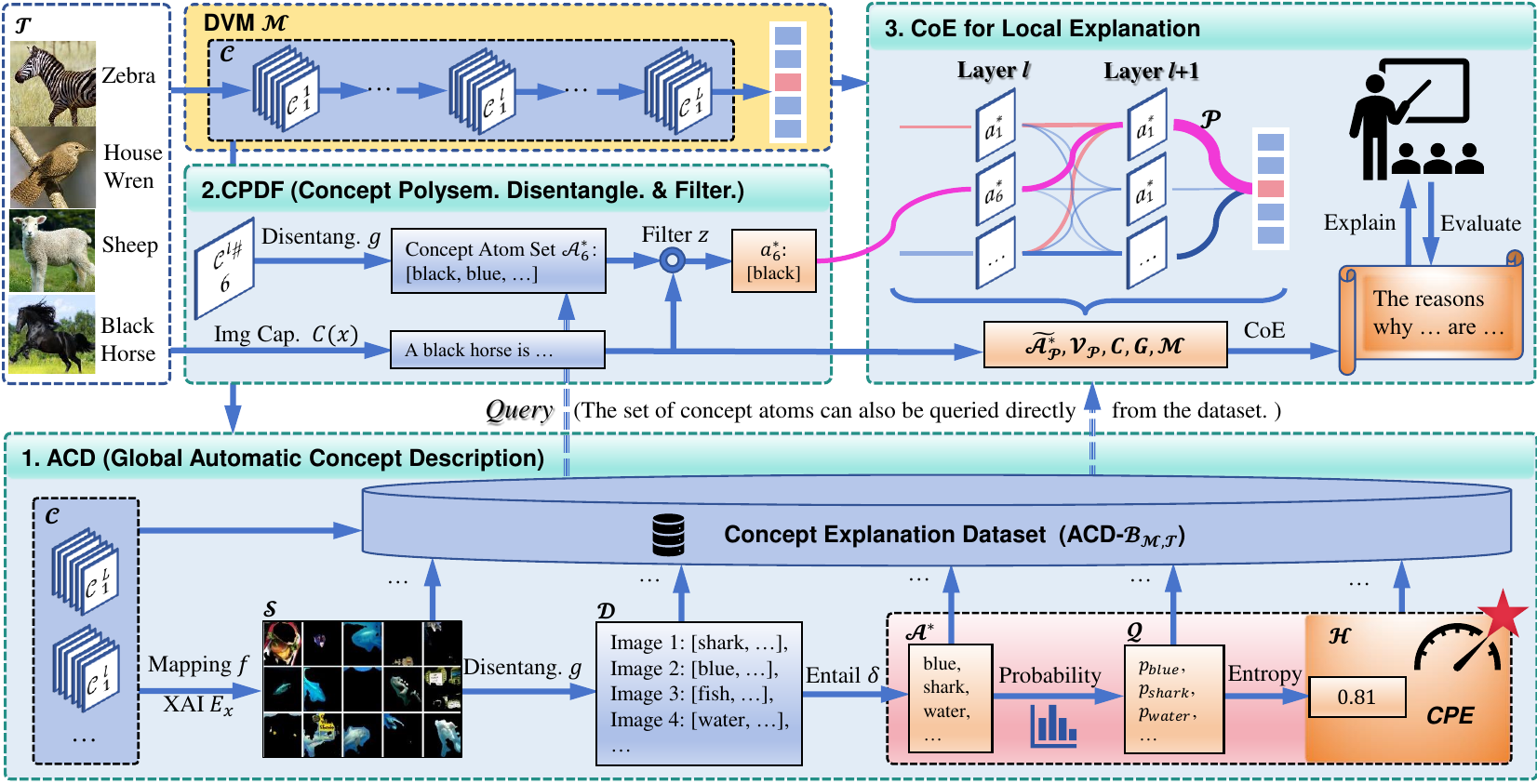}
\caption{Outline of the Chain-of-Explanation. It consists of three steps (i.e., ACD, CPDF, and local explanation). Each part utilizes a set of channels $\mathcal{C}$ from $\mathcal{M}$. In the concept circuit, $\mathcal{P}$, lines of different depths and colors represent the magnitude of the relevance.}
\vskip -0.15in
\label{fig.coe}
\end{figure*}

To alleviate this dilemma, we propose a Chain-of-Explanation (CoE) approach via automatic concept decoding, disentanglement, filtering, description, and polysemanticity quantification. 
Specifically, CoE first proposes an Automatic Concept decoding and Description (ACD) method to construct global concept explanation datasets.
It decodes every VC of key layers through arbitrary advanced XAI methods.
A Large Vision Language Model (LVLM) is utilized to describe the commonalities of the extracted image patches since the linguistic explanation is more comprehensible than images \cite{xaisurveykbs}.
As for the polysemantic concepts, we formulate a Concept Polysemanticity Disentanglement and Filtering (CPDF) mechanism to disentangle them into a set of orthogonal concept atoms, along with a Concept Polysemanticity Entropy (CPE) to quantify their degree of polysemanticity.
In this paper, 13 distinct semantic directions are prompted, including low-level and high-level semantics, which is much more than previous work (\eg, 5 in \cite{netdissect}).
Within the CPDF, a semantic entailment model is applied to mitigate the impact of redundant semantics.
After these operations, the modeling of deterministic concepts is upgraded to uncertain concept atom distributions.
When elucidating the local decision-making process, a concept circuit method is utilized to form the basis of an explanation chain, with each node representing a hierarchical set of essential concepts.
Relevant atoms and their interconnections are contextually filtered and aligned within the chain. 
Finally, we consolidate all conceptual information along the explanation chain to generate linguistic explanations similar to a Chain-of-Thought (CoT) pipeline \cite{cotsurvey1,cotsurvey4}, leveraging a Large Language Model (LLM), as shown in Fig. \ref{fig.coe}.
Qualitative and quantitative experiments validate the effectiveness and the superiority of the CoE approach.
The contributions of this paper are summarized as follows.

\begin{itemize}
\item 
We propose a novel CoE approach to automatically and systematically provide concept-based glocal explanations, along with managing the problem of polysemanticity, in a more interpretable natural language format.

\item 
A CPDF mechanism is formulated to handle the hard-to-interpret polysemantic concepts. 
To our knowledge, CPDF is a pioneer tailored to comprehensively disentangle and quantify polysemanticity.
It disentangles them into a set of succinct concept atoms, followed by a filtering function to contextually find the suitable atom as the explanation node.
A CPE score is defined to quantify the polysemanticity of concepts, which can be utilized as a metric of the interpretability of concepts or models.

\item 
Both GPT-4o and human-based experiments validate the superiority of the CoE in automatically constructing global and local explanations and the effectiveness of CPE in quantifying concept polysemanticity.

\end{itemize}

\section{Related Work}

Research on XAI can be broadly categorized into ad-hoc model interpretability and post-hoc explanation studies \cite{xaisurveytetci, xaisurveykbs}. 
The former actively refine DVMs to improve interpretability while also necessitating consideration of a trade-off between performance and interpretability \cite{xaicvprinterpre1, 2019Stop, causalcvpr, xaiinterpre2, xaitaddypami}.
The latter explains DVMs passively without performance decline \cite{explainanything, gradcam, xainankai, autoconcept,baupnas}.
Among these post-hoc efforts, concept-based explanations can provide glocal insights into DVMs \cite{lrpsurvey}.
Literature \cite{netdissect} annotated every pixel in special datasets with only 5 kinds of concepts.
The explanation is derived by analyzing the activation regions of each neuron.
MILAN \cite{milan} decoded VCs by calculating the maximum mutual information between channels and regions, followed by manually summarizing commonalities into 3 sentences. 
However, defining concepts solely through activations is insufficient.
CRP decoded VCs through the relevance value derived from the deep Taylor decomposition \cite{talor}, showing more accurate results than the activation-based methods \cite{naturecrp}.
The commonalities of the VCs were also identified manually, and one of these commonalities was selected randomly.
However, these methods encounter significant challenges, as discussed in Sec. \ref{sec.intro}.
In contrast, this paper proposes an automated concept decoding and description method based on LVLM, which offers greater flexibility in providing linguistic explanations. 
The most suitable concepts for interpretation are contextually selected, showing superior results.

Circuits, as sub-graphs, are proposed to explain the key decision route of a model \cite{circuitgpt, autocircuit, visioncircuit, fel2024holistic, distillcircuit}. 
They identify the highest-activated concepts and then explain the model's decision by linking all concepts across the circuit. 
However, its interpretability for DVMs is limited due to the lack of a cohesive, language-based explanatory output.
Additionally, the polysemanticity of VCs significantly hinders the accurate explanation \cite{understandingpoly, distillpoly}, and research in this area remains limited.
Some methods attempted to indirectly mitigate its impact by predefining principal vectors or subspaces \cite{disentanvector, fel2024holistic,tcav}.
The lack of flexible and accurate concept-level explanations also restricts their adaptability. 
Literature \cite{understandingpoly} analyzed the polysemanticity from an information compression perspective.
Pure \cite{pure} alleviated the polysemanticity by clustering and decoupling the circuit route.
These methods, however, lack the ability to simultaneously and comprehensively quantify, disentangle, and describe polysemantic VCs. Their applicability is limited—challenges that this paper seeks to address, as discussed in Sec. \ref{sec.coe-overview}.

\section{Chain-of-Explanation Approach}
\label{sec.coe-overview}

In this paper, a CoE approach is proposed to automatically explain DVMs from both global and local perspectives. 
It breaks down the complex reasoning path into a chain of concept nodes, within which polysemanticity is effectively managed.
As shown in Fig. \ref{fig.coe}, CoE is composed of three steps. 
\textbf{Step 1:} ACD is proposed to automatically construct global concept descriptive explanation datasets of a given DVM.
\textbf{Step 2:} With the assistance of an LVLM, a CPDF mechanism is designed to disentangle each original VC into a set of concept atoms and extract the most suitable atom according to the context, making CoE more comprehensible and flexible. 
Besides, a CPE score is formulated accordingly to quantify the degree of polysemanticity of concepts.
\textbf{Step 3:} A local explanation chain in a manner akin to a CoT for the decision-making progress of the DVM is established based on the dataset, CPDF, concept circuit, and LLMs.

\subsection{Automatic Concept Decoding and Description}

ACD method is first formulated to automatically and effectively construct databases of global concept descriptions.
Specifically, given an image dataset $\mathcal{T}$ and a DVM $\mathcal{M}$ requiring evaluation and explanation, we establish an injective function $f$ between channel set $\mathcal{C}$ and visual semantic concept set $\mathcal{S}$ by an XAI method $E_x$, $f: \mathcal{C} \mapsto \mathcal{S}.$
Without loss of generality, for $j_{th}$ channel $c_j^l$ in layer $l$, $E_x$ extracts a set of image patches within $\mathcal{T}$ that most activate this channel, named VC $s_j^l = \left\{ I_n \right\}_N$, where $N$ is the number of image patches. 
Each $I_n$ highlights certain regions of the original image, representing the focused concept, as exemplified in the bottom left part of Fig. \ref{fig.coe}.

A describing function $r$ is then defined to decipher $\left\{ I_n \right\}_N$ into a set of language descriptions:
\begin{equation} \label{injective}
  \mathcal{D} = r (\left\{ I_n \right\}_N) = \left\{ d_m \right\}_M,
\end{equation}
where $M$ is the number of language descriptions for the current channel.
Traditionally, $r$ is implemented by manually and subjectively identifying commonalities within $\left\{ I_n \right\}_N$ \cite{netdissect}, while ACD utilizes a powerful LVLM as the describer with well-engineered prompts.
After constructing $\mathcal{S}$, $\mathcal{D}$ and the mapping function $f$, a concept description dataset $\mathcal{B}_{\mathcal{M},\mathcal{T}} = \left\{\mathcal{C}, \mathcal{S}, \mathcal{D}, f \right\}$ is automatically constructed, providing global linguistic explanations for all VCs of $\mathcal{M}$.

\subsection{Managing Polysemantic Concept}

VCs of each node within a DVM exhibit polysemanticity, introducing certain deviations in the explanation framework, as shown in Fig. \ref{fig.coe}.
Higher polysemanticity indicates that the current VC aggregates more semantics, making the intended representation more ambiguous. 
Such semantic uncertainty impedes human comprehension of the concept, particularly within the CoE, where various concepts must be understood collectively.
Quantifying and solving this dilemma is difficult and has been underestimated by previous studies.
Under this circumstance, we propose a CPDF mechanism along with a CPE score to mitigate and quantify its detriments to the CoE simultaneously.

\subsubsection{Polysemantic Concept Decoupling and Filtering}

Considering that a polysemous word can be understood as an aggregation of various distinct meanings, we define a function $g$ to disentangle the ambiguous VC into a set of easily understandable linguistic concept atoms:
\begin{equation} \label{eq.disnaive}
   \mathcal{A} = \left\{a^1,\cdots,a^i,\cdots,a^Q \right\}=g(s),
\end{equation}
where $Q$ is the number of the concept atoms.
We posit that each atom represents a view of the current VC $s$, which will be activated upon the appearance of the corresponding semantic in the input.
Furthermore, given that concept atoms should accurately represent the commonalities in $\left\{ I_n \right\}_N$ and recognizing that this naive disentanglement function (Eq. \ref{eq.disnaive}) does not provide a foundation for quantifying polysemanticity, we refine the function $g$
to disentangle each image patch $I_n$ into a set of atoms:
\begin{equation} \label{eq.dis}
\begin{gathered}
    \mathcal{A}_n = \left\{a^1_n,\cdots,a^i_n,\cdots,a^Q_n \right\}=g(I_n, s),\\
    \mathcal{D} = \left\{\mathcal{A}_n\right\}_N.
\end{gathered}
\end{equation}

We instantiate $g$ by a powerful LVLM (\eg, GPT-4o \cite{gpt4o}), considering that this is quite an open and complex problem. 
$g$ prioritizes disentangling and describing commonalities within the VC while analyzing each individual image.
Various commonalities among any certain number of image patches should be summarized and described by precise terminologies.
These common semantics within an open scenario are as diverse as they are abundant, unfolding in countless forms and nuances.
Besides, it should generate specific atoms when any commonalities identified above do not appear in the current $I_n$.
Therefore, the art of prompt engineering \texttt{prompt-com} becomes paramount.
It should provide rules for recognizing commonalities, possible conceptual directions, and output format control. 
In this paper, we prompt 13 high-level and low-level semantical directions that cover the normal concepts encountered in daily life, including object category, scene, object part, color, texture, material, position, transparency, brightness, shape, size, edges, and their relationships.
Additionally, the CoT techniques \cite{cotsurvey1, cotsurvey4} are introduced.
Details of all sophisticated prompts are shown in Sec. S2 of the Appendix.

All $\mathcal{A}_n$ are then aggregated into a description atom set $\mathcal{A}$ with all duplicate atoms removed:
\begin{equation} \label{eq.Aset}
\begin{aligned}
 &\mathcal{A}&= \left\{\mathcal{A}_1 \cup \mathcal{A}_2 \cup \cdots \cup \mathcal{A}_N \right\} \\
 &&= \left\{a^1,\cdots,a^i,\cdots,a^P \right\},
\end{aligned}
\end{equation}
where $P$ is the number of concept atoms after deleting the repeating atoms.
Each element in $\mathcal{A}$ represents a disentangled monosemantic concept and can be utilized as a node of the CoE.
However, some atoms are semantically equivalent even if they have variant descriptions \cite{nature_semanticentropy}.
In order to remove semantically redundant atoms, we introduce an entailment function $\delta$ to detect the relations between atoms:
\begin{equation} \label{eq.entail}
  \left\{ 1,0,-1 \right\} = \delta(a^{i1}, a^{i2}), \   \forall a^{i1}, a^{i2} \in \mathcal{A},
\end{equation}
where the responses 1,0,-1 represent entailment, neutral, or contradiction of the candidates, respectively. 
We instantiate $\delta$ by an NLI model (\eg, DeBERTa model \cite{deberta}) because the atom candidates are relatively short and simple that an NLI model is good enough. 
Following \cite{nature_semanticentropy}, a bidirectional entailment check is applied, meaning that one of the candidates will be removed when the response is entailment or neutral.
The first atom is designated as the monosemantic representative of the semantic equivalence set due to the transitivity between them.
The final concept set $\mathcal{A}^\ast$ only contains heterogeneous atoms that describe the current VC:
\begin{equation} \label{eq:Astar}
  \mathcal{A}^\ast = \left\{a^1,\cdots,a^i,\cdots,a^{P^\ast} \right\}.
\end{equation}

After disentangling the concept atoms, we define a filtering function $z$ to select an atom $a^\ast$ as the most suitable explanation of current VC with respect to the input $x$:
\begin{equation} \label{eq.filter}
  a_s^\ast(x) = z(\mathcal{A}^\ast_s, C(x)).
\end{equation}
Some research ignores this step or selects the concept randomly with an implicit assumption that all concepts in $\mathcal{A}^\ast$ are semantically equivalent \cite{milan, devil}.
In contrast, we filter the most suitable atom $a^\ast$ based on the context of the CoE. 
In particular, $z$ is instantiated by an LLM. 
$C(x)$ is an image captioning function utilized for providing contextual information. 
It can be instantiated by a relatively small LVLM \cite{intern1, intern2} or a professional model \cite{capanything, blip2}.

\subsubsection{Quantification of Concept Polysemanticity}

The polysemanticity of concepts can serve as a metric for the model's interpretability. 
Lower polysemanticity of concepts indicates reduced semantic uncertainty with a more singular form of activation, leading to greater interpretability.
In this paper, we endeavor to quantify polysemanticity from three levels of granularity (i.e., conceptual channels, layers, and models) by defining a CPE score. 
The larger the CPE, the more pronounced the polysemanticity, which leads to decreased interpretability.

In particular, the disentanglement function $g$ in Eq. \ref{eq.dis} enables us to count the frequency of each atom in $\mathcal{A}^\ast$. 
Given two fixed numbers $Q$ and $N$, we calculate the probability of the $i_{th}$ atom within $j_{th}$ concept:
\begin{equation} \label{eq.prob}
    p_i = \frac{Num_i}{Q \times N +Pad}.
\end{equation}

Then, the CPE of $j_{th}$ concept can be formulated as
\begin{equation} \label{eq.Hjstar}
    H_j^\ast = \frac{-\sum_{i=1}^{P^\ast + Pad} p_i \log p_i}{\log (P^\ast+Pad)}.
\end{equation}
The denominator normalizes the entropy value to a range between 0 and 1, considering that the size of $P^\ast$ varies for different VCs.
Besides, the quantity of $\mathcal{A}^\ast$ is set to be at least equal to $N$ due to some $\mathcal{D}_s$ having a small number of semantical equivalent atoms with equal probabilities. 
If there are insufficient unique atoms, we supplement the corresponding number of atoms, with each supplementary atom set to a number of 1. 
The atomic objects are expanded, and the denominator is increased by the number of supplemented atoms (a total amount of $Pad$). 
Please refer to Sec. S3.1 of the Appendix for details.
In addition, two averaged CPEs are formulated to enable comparisons of polysemanticity between specific layers or models:
\begin{equation} \label{eq.Hjlayer}
    H_l^\ast = \frac{1}{d_l} \sum^{d_l} H_j^\ast,
\end{equation}
\begin{equation} \label{eq.Hjmodel}
    H_{\mathcal{M}}^\ast = \frac{1}{L} \sum^{L} H_l^\ast,
\end{equation}
where $d_l$ is the dimension of the channel and $L$ is the number of layers.
Finally, to accommodate the diverse needs of researchers within the community, the automatically constructed concept explanation dataset is extended as ACD- $\mathcal{B}_{\mathcal{M},\mathcal{T}} = \left\{\mathcal{C}, \mathcal{S}, \mathcal{D}, \mathcal{A^\ast}, \mathcal{Q}, \mathcal{H}, f \right\}$, where $\mathcal{Q}$ represents the probability set of each atom and $\mathcal{H}$ is the set of CPE.

\subsection{Construction of Explanation Chain}

Concept circuits in DVMs can identify the most activated units in each layer for a specific input \cite{visioncircuit}. 
When aggregating these units across all key layers in a bottom-up manner, a corresponding maximum activation path $\mathcal{P}$ is generated, establishing the foundation for the explanation chain.
Each node in this path serves as the explanation for the model’s decision at that layer. 
Traditional concept circuit methods for DVMs identify the activation path without providing thorough overarching natural language-based explanations. 
Inspired by the CoT explanation technique in LLMs \cite{cotsurvey2}, CoE proposes a local explanation chain to interpret the decision-making process of DVMs in the form of natural language descriptions based on ACD-$\mathcal{B}_{\mathcal{M},\mathcal{T}}$, CPDF, concept circuit method, and LLM, as shown in Fig. \ref{fig.coe}.

Specifically, some user-predefined layers requiring explanations or key layers facilitating CoE (\eg, output layers of 4 stages of a DVM) are listed in a bottom-up manner.
Given an image $x$ and layer $l$, an advanced XAI method $E_x$ computes the maximization normalized activation or relevance values for all concepts. 
Each value represents the importance of the current concept to the prediction $\mathcal{M}(x)$.
CoE acknowledges top $k^l$ relevant concepts 
$ \mathcal{C}^l(x) = \left\{ c_1^l, \cdots, c_{k^l}^l \right\} $ 
and their values 
$ \mathcal{V}^l(x) = \left\{ v_1^l, \cdots, v_{k^l}^l \right\} $, determined by the $\alpha$ quantile of relevance values across all concepts
(i.e., $k^l = | \left\{v_j^l | v_j^l \textgreater \alpha \max (v^l), \forall j \in d^l \right\}|$, 
where $|\mathcal{X}|$ computes the number of elements).
After that, the descriptive concept atoms of the current layer $ \widetilde{\mathcal{A}}^{\ast, l}_{k^l}(x) = \left\{ a_1^{\ast l}, \cdots, a_{k^l}^{\ast l} \right\} $ are constructed based on ACD-$\mathcal{B}_{\mathcal{M},\mathcal{T}}$ and the filter Eq. \ref{eq.filter}, serving as a node of the explanation chain.

The final CoE for the local decision-making explanation, abstracting and synthesizing all top activated concept atoms along the chain, is formulated as
\begin{equation} \label{eq.coelocal}
\hspace{-0.8mm}
\text{\fontsize{9.6}{10}\selectfont {$CoE_{\mathcal{M}}(x) = e ({\textstyle \widetilde{\mathcal{A}}^{\ast}_\mathcal{P}(x), \mathcal{V}_\mathcal{P}(x), C(x), G(x), \mathcal{M}(x)})$}},
\end{equation}
where $e$ formulates a synthesizing and describing function with inputs obtained from the above sections, and $G(x)$ is the label of $x$. 
We instantiate $CoE_{\mathcal{M}}(x)$ through a powerful LLM (\eg, GPT-4 \cite{gpt4}), with prompt \texttt{prompt-coe} well-designed by combining CoT and few-shot prompting \cite{promptsurvey}.

Additionally, these local explanations are evaluated by a powerful LVLM and human judges based on three explainability metrics, each utilizing a three-tier scoring system.

\begin{table*}[!h]
\caption{Examples of VCs and their disentangled concept atoms, along with their concept probability distributions and CPE scores. The first row shows concept 75 of Layer 4 extracted by CRP. The second row shows concept 697 of Layer 4 extracted by MILAN.}
\label{tab.egmain}%
\centering
\begin{tabular}{m{11cm}<{\centering}m{4.2cm}<{\centering}|m{0.8cm}<{\centering}}
\toprule
    Visual Concept and Disentangled Concept Atoms & Concept Distribution & CPE \\
\toprule
   \includegraphics[width=1\linewidth]{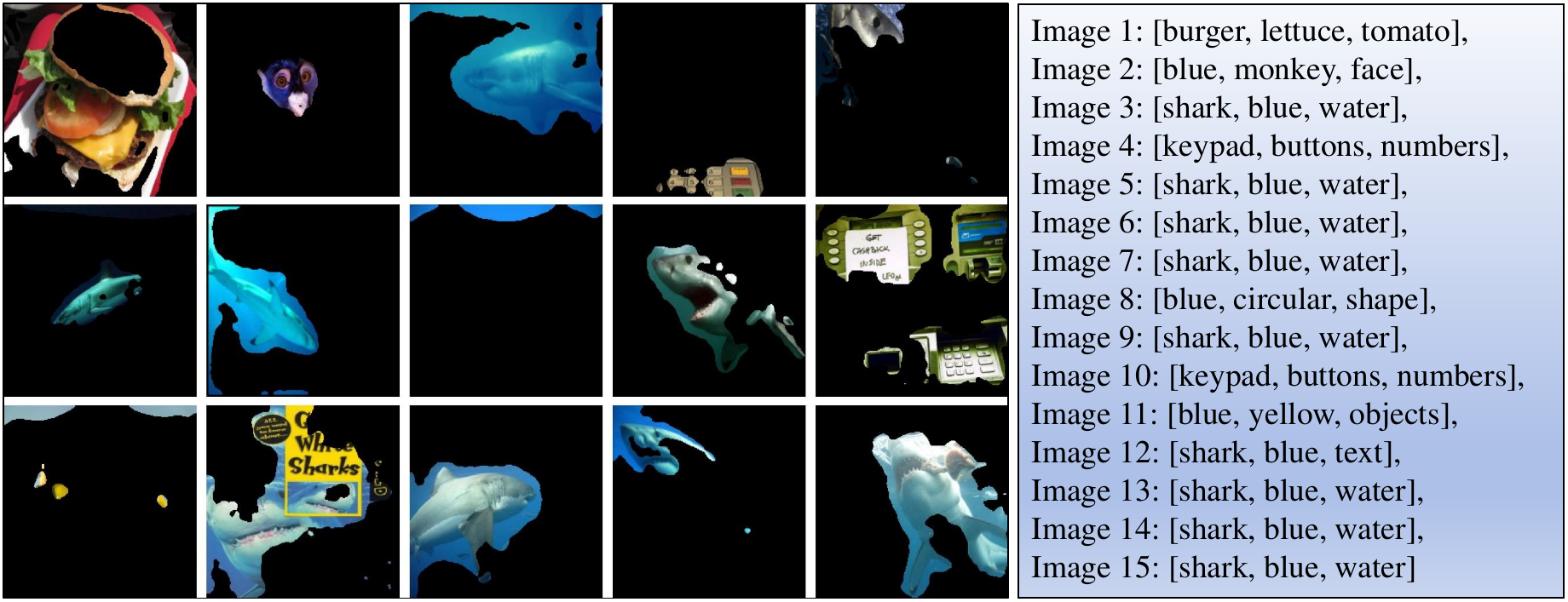} 
   & \includegraphics[width=1\linewidth]{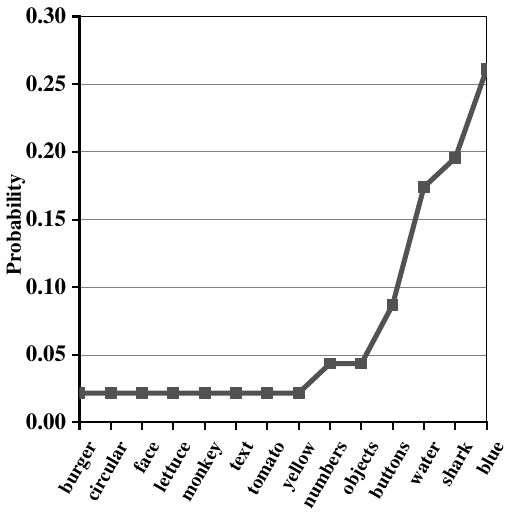}
   &\multirow{1}{*}{\textbf{0.81}}
   \\
\midrule
   \includegraphics[width=1\linewidth]{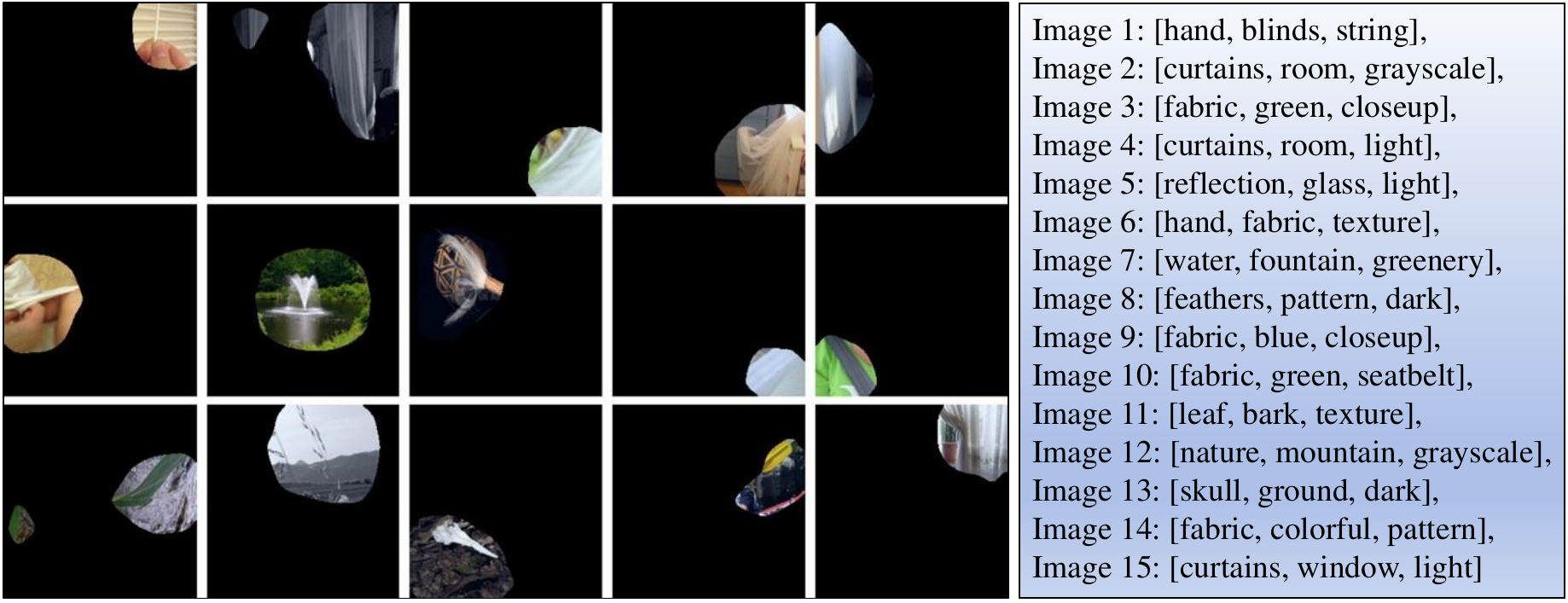} 
   & \includegraphics[width=1\linewidth]{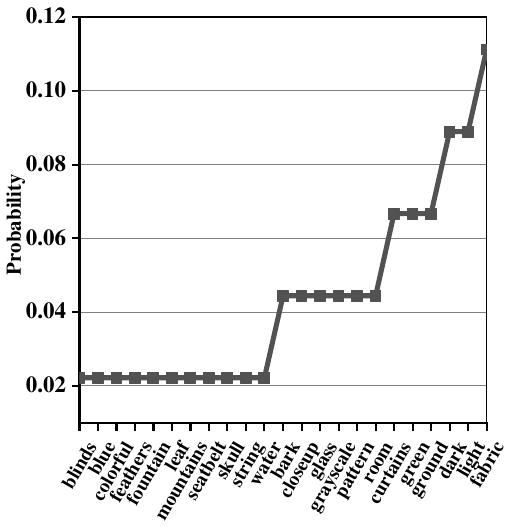}
   &\multirow{1}{*}{\textbf{0.95}}
   \\
\bottomrule
\end{tabular}%
\vspace{-0.15in}
\end{table*}%

\section{Experiments}

We evaluate the proposed CoE approach on two types of DVMs (i.e., ResNet \cite{resnet}
and CLIP \cite{clip}). 
The relevance-based XAI method CRP is adopted as the primary approach for $E_x$ in this paper, owing to its demonstrated superiority in terms of fidelity and reliability  \cite{naturecrp}.
Please refer to the Appendix for more details on the implementation and results.
Codes are available at \href{https://github.com/YuWLong666/CoE}{https://github.com/YuWLong666/CoE}.

\subsection{Concept Explanation Dataset Construction}

In the CoE, we first automatically construct global concept description databases ACD-$\mathcal{B}_{\mathcal{M}, \mathcal{T}}$ for various DVMs.
Taking a ResNet152 model as an example, 4 output layers of 4 stages (with 256, 512, 1024, and 2048 channels $\mathcal{C}$, respectively) are selected as the key layer set.
CRP is applied as $E_x$ to extract the VC set $\mathcal{S}$.
Following \cite{naturecrp}, each VC is represented by 15 (i.e., $N$) image patches. 
The mapping function $f$ from channels to VCs is built.
We then employ a GPT-4o-2024-08-06 model as the function $g$ to generate the description set $\mathcal{D}$ (i.e., Eq. \ref{eq.dis}).
The disentangled and clustered atom set $\mathcal{A^\ast}$ (processed by a DeBERTa model), the probability set $\mathcal{Q}$, and the CPE set $\mathcal{H}$ are computed and gathered accordingly.
Overall, ACD-$\mathcal{B}_{\mathcal{M}, \mathcal{T}}$ containing 3,840 entries is constructed, with each entry comprising a channel, a CPE score, and 15 VC representations—each consisting of 3 concept atoms (i.e., $Q$)—and multiple orthogonal linguistic atoms, along with their associated probabilities.

\begin{figure*}[t]
\centering
\begin{subfigure}{0.33\linewidth}
    \centering
    \includegraphics[width=0.98\columnwidth]{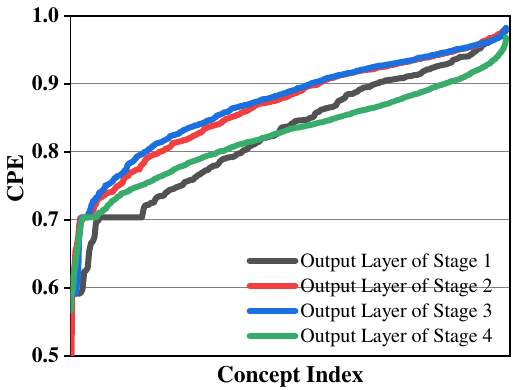}
    \caption{Sorted CPE along the channel dimension.}
    \label{fig.egcpe.concept}
\end{subfigure}
\begin{subfigure}{0.33\linewidth}
    \centering
    \includegraphics[width=1\columnwidth]{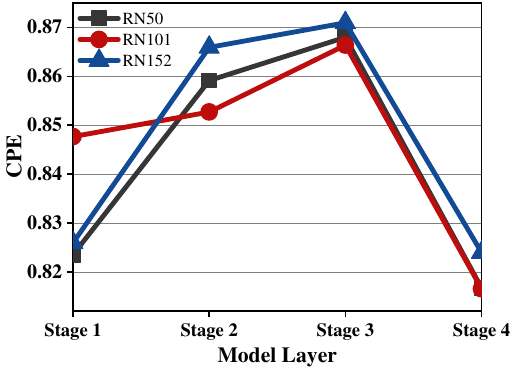}
    \caption{Averaged CPE on 4 key layers.}
    \label{fig.egcpe.layer}
\end{subfigure}
\begin{subfigure}{0.33\linewidth}
    \centering
    \includegraphics[width=0.95\columnwidth]{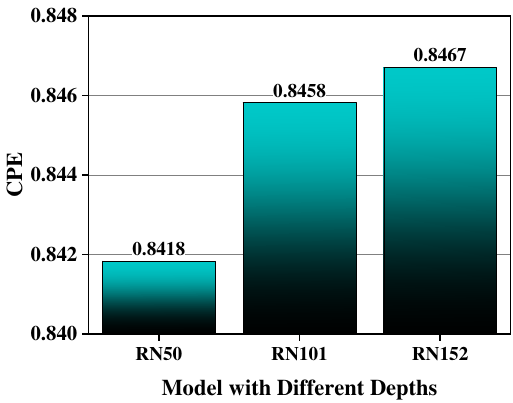}
    \caption{Averaged CPE on 3 models.}
    \label{fig.egcpe.cnnmodel}
\end{subfigure}
\caption{Illustrations of CPE scores at 3 levels of granularity. In (a), the channel indices of each layer are max normalized to 1.
}
\vskip -0.15in
\label{fig.egcpe}
\end{figure*}

\subsection{Disentanglement of Polysemantic Concept}

CPDF automatically disentangles and describes the polysemantic VCs into a set of linguistic concept atoms, $\mathcal{D}$.
As exemplified by Table \ref{tab.egmain}, two randomly selected VCs are extracted by two advanced XAI methods (CRP and MILAN).
They are disentangled and described into atoms, exhibiting different degrees of polysemanticity.
We observe that the commonalities within the VCs are almost entirely identified, and most of the atoms describe the commonalities of subsets of image patches.
In the first row (CPE=0.81), $g$ successfully identifies all visually discernible commonalities, such as shark and water.
The atoms exhibit high consistency, and fewer but more frequently occurring common semantics are identified.
Only a few atoms, such as the burger, are not commonalities, as their corresponding images lack sufficient common semantics.
The atoms are also represented accurately. 
For instance, the atom sets for each image featuring the shark semantic all include the word shark. 
In the second row (CPE=0.95), our method remains effective in disentangling and describing VC with a higher level of polysemanticity.  
$g$ identifies a greater variety of atoms.
These results demonstrate the effectiveness of the disentanglement and description operation in the CPDF.

\subsection{Concept Distribution and CPE}

CPE can quantify the polysemanticity and serve as an indicator of a DVM's interpretability. 
It begins by calculating the probabilities of each orthogonal atom. 
As the middle column of Table \ref{tab.egmain} shows, the distribution provides statistics on the extent to which different atoms are associated with each concept.
For instance, concept 75 primarily focuses on blue with a probability of 0.27 while also identifying yellow with a smaller probability.
The semantics between atoms are non-overlapping.
This probabilistic approach enhances explainability by capturing multiple semantics with their occurring probabilities, upgrading the modeling of deterministic concepts to uncertain atom distributions.

CPE quantifies polysemanticity from 3 levels of granularity. 
As shown in Fig. \ref{fig.egcpe.concept}, 
concepts exhibit varying degrees of polysemanticity, with considerable fluctuations (ranging from a maximum of 1 to a minimum of 0.5). 
Most of the concepts demonstrate a noticeable level of polysemanticity.
When analyzing the polysemanticity of different layers, as shown in Fig. \ref{fig.egcpe.layer}, the averaged CPE peaks in the third stage, while the shallowest and deepest layers exhibit lower values. 
Typically, deeper layers capture abstract concepts, whereas the shallower layers extract elementary features like textures \cite{netdissect, tip}. 
DVMs show better monosemanticity when identifying elementary and abstract concepts, while more ambiguous concepts are captured in the intermediate layers. 
This experiment validates the theoretical and experimental claims of previous work \cite{icml2019best, devil}.
In Fig. \ref{fig.egcpe.cnnmodel}, we observe that as the number of layers increases, the CPE of the model shows a rising trend.
This result indicates a direct proportional relationship between the polysemanticity and the complexity of DVMs, which is that the more complex a DVM becomes, the harder it is to interpret.

\begin{table}[t]
    \caption{The consistency of the recognition of concept polysemanticity between the CPE metric and human scorers.}
    \label{tab.cpeconsency}
    \vspace{-0.09in}
    \centering
    \setlength{\tabcolsep}{4.5mm}{
    \begin{tabular}{c|c}
    \toprule
      & Consistency \\
    \midrule
    CPE Metric and Human Scorers   &   75 \% \\
    \bottomrule
\end{tabular}  
}
\vspace{-0.15in}
\end{table}

The CPE is evaluated in comparison with human assessments.
As shown in Table \ref{tab.egmain}, the CPE is consistent with the degree of polysemanticity from the human intuitive observations (the polysemanticity of the first row is smaller than that of the second row).
Similarly, we randomly sample 300 pairs of VCs, each containing two VCs with varying levels of polysemanticity. 
Comparisons are drawn from these 300 pairs by both human evaluators and the CPE metric.
As shown in Table \ref{tab.cpeconsency}, results from human assessments and CPE achieve a 75\% agreement rate, demonstrating the effectiveness of CPE.
Other results are shown in Sec. S4 of the Appendix.
All of these results reveal that CPE is effective in quantifying the polysemanticity of concepts and DVMs.

\begin{table*}[!t]
\caption{Cases of local explanations of CoE. The first row shows the explanation of a correct prediction, and the second row is reversed.}
\label{tab.coelocal}%
\centering
\setlength{\tabcolsep}{1mm}{
\begin{tabular}{c}
\toprule
    \includegraphics[width=0.99\linewidth]{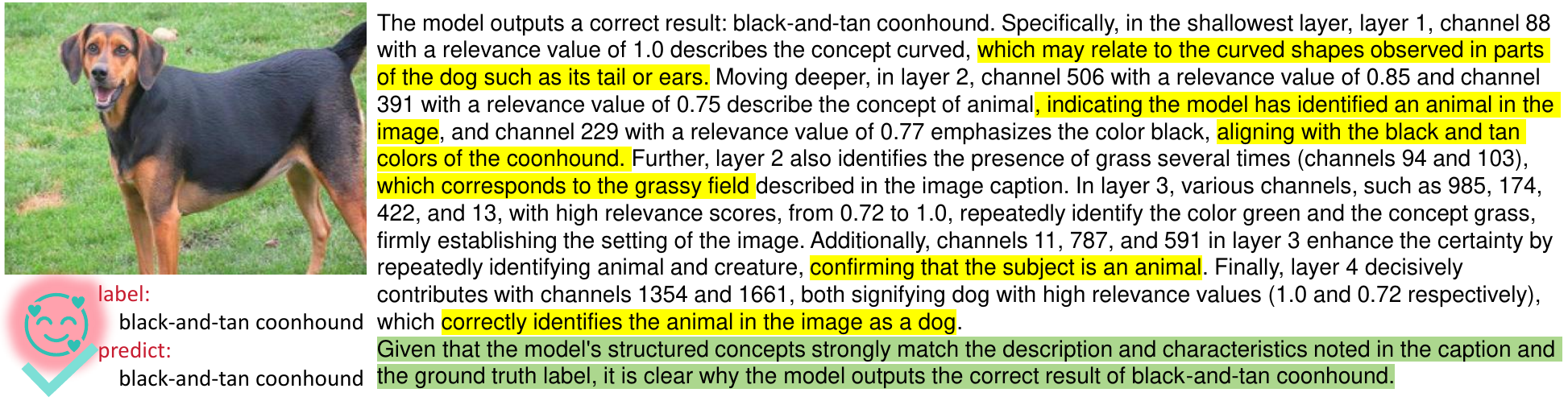}\\
    \midrule
    \includegraphics[width=0.99\linewidth]{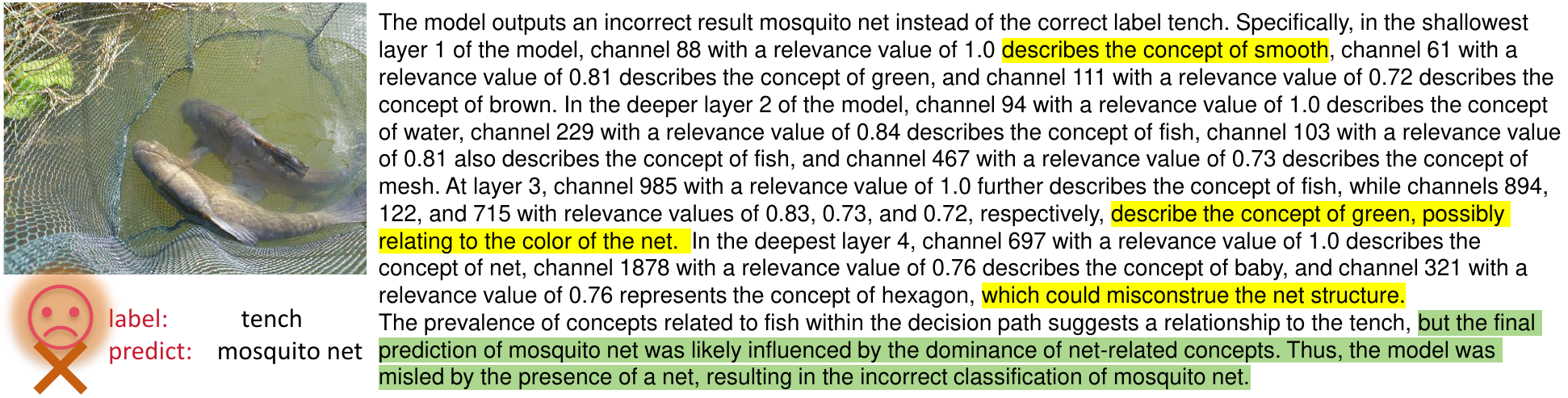}\\
\bottomrule
\end{tabular}%
\vspace{-0.06in}
}
\end{table*}%

\subsection{CoE for Local Explanations}

CoE generates local explanations akin to CoT by automatically describing the concept circuit.
We set $\alpha = 0.001$ to acknowledge the top relevant VCs, resulting in an average of 3 key VCs per layer.
As exemplified in Table \ref{tab.coelocal}, given an input image, CoE first outputs whether the prediction of the DVM is correct and then gives reasons behind this prediction according to the information along the concept circuits.
CoE employs an LLM as an engine to infer the logical relationships among concepts, as well as between concepts and the context. 
It synthesizes all evidence to generate an inductive commentary on the predictions. 
The explanation chain is presented in a linguistic format, enhancing human understanding.
For instance, CoE explains why the DVM incorrectly predicted a tench as a mosquito net: the top activation of mosquito-net-related concepts, rather than those related to fish, misled the decision.
Additional instances are shown in Sec. S5 of the Appendix.
Overall, CoE effectively captures and elucidates the core decision-making evidence for both correct and incorrect predictions of DVMs.

\begin{table*}[t]
    \caption{Scores of CoE for the local explanation on three methods evaluated by GPT-4o and human scorers. The first method utilizes MILAN ANNOTATION as the concept explanation dataset, while the rest two work on the ACD-$\mathcal{B}_{\mathcal{M}, \mathcal{T}}$ datasets.
    \textbf{Each criterion has a maximum score of \underline{2} points. The total explanation score is \underline{6} points.} Higher scores indicate better explainability.}
    \label{tab.expscore}
    \centering
    \begin{tabular}{c|l|c|c|c|c}
    \toprule
    & Method  & Accuracy & Completeness & User Interpret. & Total Explanation \\
    \midrule
    \multirow{3}{*}{GPT-4o} 
    & MILAN A.\cite{milan} + Description (Baseline) & 1.03 & 1.07 & 1.04 & 3.14 \\
    & CoE w/o Filtering     & 1.07 & 1.10 & 1.08 &3.25 \\
    \rowcolor{c1!20}
    \cellcolor{white} & CoE (ours)      & \textbf{1.69} & \textbf{1.70} & \textbf{1.67} &\textbf{5.06} \\
    \midrule
    \multirow{3}{*}{Human} 
    & MILAN A.\cite{milan} + Description (Baseline) & 0.87 & 0.83 & 0.92 & 2.62 \\
    & CoE w/o Filtering    & 1.10 & 1.16 & 1.07 & 3.34 \\
    \rowcolor{c1!20}
    \cellcolor{white} & CoE (ours)           & \textbf{1.73} & \textbf{1.72} & \textbf{1.59} &\textbf{5.04} \\
\bottomrule
\end{tabular}            
\vspace{-0.08in}
\end{table*}

\subsection{Quantitative Evaluation via GPT-4o \& Human}

We employ GPT-4o and human evaluators to assess the local explanations for 500 and 100 samples randomly selected from the ImageNet Validation dataset \cite{imagenet}, respectively. 
The GPT-4o evaluation is achieved through a carefully designed prompt \texttt{prompt-coe-eval}.
Explanations generated by three methods are rated according to three explainability metrics: Accuracy, Completeness, and User Interpretability \cite{xaievaluation}.
Each is assigned a three-tier scoring system (2, 1, and 0), with specific criteria defined for each tier, resulting in a total explanation score of 6.
Please refer to Sec. S2 and S5 of the Appendix for details.

As shown in Table \ref{tab.expscore}, our CoE approach attains peak performance across all three metrics, achieving total scores of 5.06 and 5.04 in GPT-4o and human evaluations, respectively.
The lowest scores (3.14 and 2.62) are obtained on the MILAN ANNOTATION dataset without the ACD and CPDF methods.
This dataset was manually annotated, resulting in inaccuracy and insufficiency within concept explanations.
Scores increase by 0.11 and 0.72, respectively, with the application of our proposed ACD-$\mathcal{B}_{\mathcal{M}, \mathcal{T}}$ dataset.
The ACD method not only enables the automated disentanglement and description of concept atoms but also enhances their precision and comprehensibility.
When the CoE employs the full CPDF mechanism, the explanation score rises by 1.92 and 2.42 points (the average absolute improvement is 36\%), respectively, with the filtering step contributing 1.81 and 1.70 points.
This result suggests that by taking polysemanticity and contextual filtering of concepts into account, CoE effectively guides the LLM in producing better linguistic explanations, substantially enhancing their accuracy, completeness, and interpretability for users.
The results of the human evaluation are consistent with the GPT-4o results.
All of these results strongly affirm the superiority of the proposed CoE approach.

\section{Conclusion}

In this paper, we propose a CoE approach to tackle the challenges of the inflexibility to automatically construct concept explanation datasets, insufficient linguistic explanations, and the weakness in managing concept polysemanticity of concept-based XAI methods.
CoE automates the decoding and description of concepts via an LVLM and constructs global concept explanation datasets. 
A CPDF mechanism is designed to identify the most contextually relevant concept atoms, advancing the modeling of deterministic concepts to uncertain concept atom distributions.
The CPE is proposed to quantify the polysemanticity.
Finally, CoE enables local explanations of a DVM’s decision-making process, represented through natural language. 
The effectiveness and superiority of CPE and CoE are demonstrated.
In summary, CoE establishes a systematical framework for explaining DVMs, and we believe that the findings presented in this paper warrant further exploration.

\clearpage  

\section*{Acknowledgment}
This research is supported in part by the National Natural Science Foundation of China under Grants U23B2049, 62276186, and 61925602.

{
    \small
    \bibliographystyle{ieeenat_fullname}
    \bibliography{main}
}

\input{X_suppl}

\end{document}

%% file: X_suppl.tex
\clearpage
\setcounter{page}{1}
\maketitlesupplementary
\renewcommand{\thetable}{S\arabic{table}}
\renewcommand{\thefigure}{S\arabic{figure}}
\renewcommand{\thesection}{S\arabic{section}}

\addtocontents{toc}{\protect\setcounter{tocdepth}{2}}
\setlength{\cftsecnumwidth}{2.1em}
\setlength{\cftsubsecnumwidth}{2.8em}
\setcounter{section}{0}

\tableofcontents

\section{Overview of Supplementary Material}

In this supplementary material, we mainly give detailed information and analyses about prompt engineering, the CoE approach, and experimental results of the CPE and CoE local explanations.
Specifically, as presented in Sec. \ref{sec.prompt}, the prompts that are well-designed in this paper include \texttt{prompt-com} for automatically disentangling and describing the commonalities of the given VCs, \texttt{prompt-coe} for generating the local explanations given specific samples, and \texttt{prompt-coe-eval} for evaluating the local explanations from three explainability metrics.
In Sec. \ref{sec.coeapproach}, we present details of the CoE approach, highlighting its distinguishing qualities.
We also discuss the distinctions between CoE and CoT.
In Sec. \ref{sec.sup.exp-cpe}, we give experimental results of various versions of CPE, including naive CPE, CPE with clustering, and our final refined version.
We also analyze the CPE results explored from different XAI methods (i.e., relevance, activation, and maximum mutual information-based methods) and different model architectures (i.e., ResNet and CLIP).
Various examples of VCs, their disentangled concept atoms, probabilities, and CPE values are provided. 
Besides, we give details on the implementation of the comparison experiments between human evaluations and the CPE metric.
In the Section \ref{sec.sup.exp-coe}, we provide details of the evaluation criteria for linguistic local explanations, along with an overview of the human evaluation process.
Details of the comparison between local linguistic explanations generated by different methods and various instances of these local explanations are presented.

\begin{table}[t]
\caption{The prompt template of \texttt{prompt-com}. 
Before querying the LVLMs, we substitute the curly brackets with actual texts.}
\label{tab.appe-prompt-com}%
\centering
   \includegraphics[width=0.91\columnwidth]{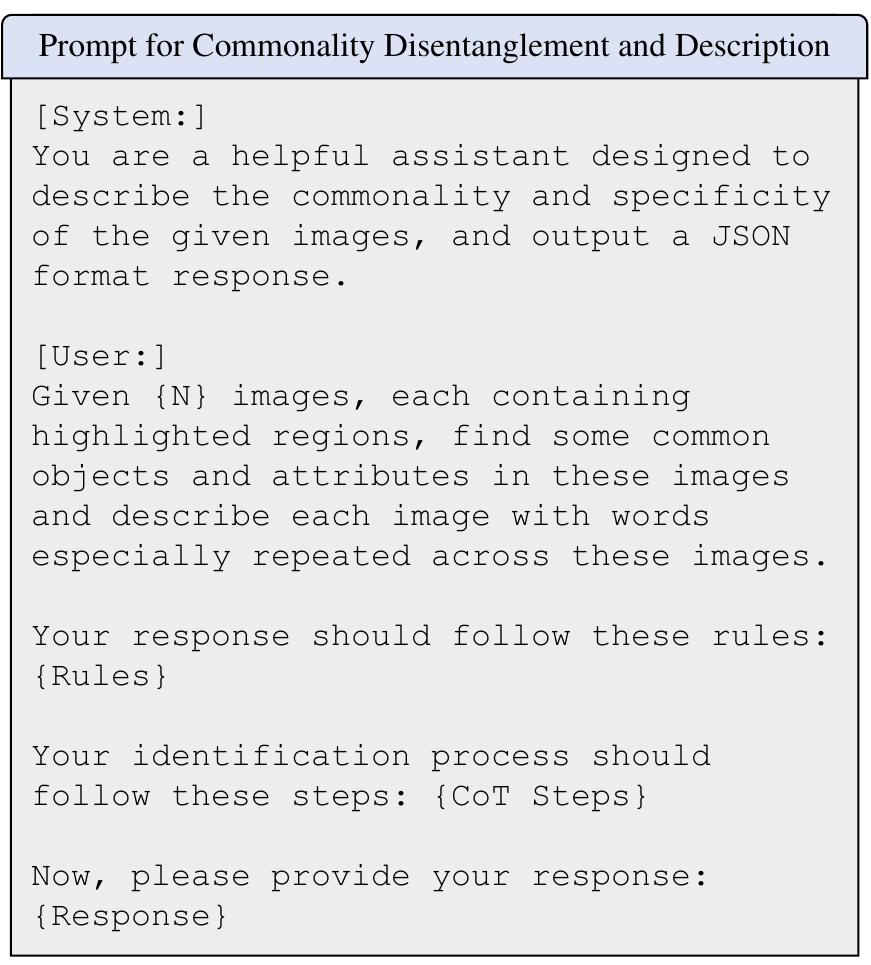} 
\vskip -0.1in
\end{table}

\section{Prompt Engineering}
\label{sec.prompt}

In this section, we provide a detailed exposition of the three well-engineered prompts designed to describe commonalities of VCs, aggregate all information along the concept circuit to enable the CoE to generate local explanations and evaluate the generated local explanations.

\subsection{Prompt for Commonality Describing}

In this paper, we design a sophisticated prompt \texttt{prompt-com} to describe the commonalities by a set of concept atoms. The meticulously crafted prompt template is presented in Table \ref{tab.appe-prompt-com}.
As discussed in the main manuscript, this prompt is engineered to accurately disentangle and summarize the commonalities across multiple subsets of images utilizing precise terminology drawn from 13 semantic directions.
Additionally, the disentangled atoms also serve as the foundation for the probability and CPE calculations.
To meet these requirements, this prompt incorporates some rules along with step-by-step guidance.
The rules outlined below primarily establish 13 semantic directions and delineate the format for output control.

\begin{ttfamily}

1. Pay more attention to the repeated objects or attributes across these images.

2. Possible objects or attributes you can use to describe these images are object category, scene, object part, color, texture, material, position, transparency, brightness, shape, size, edges, and their relationships.

3. The identified common objects or attributes must appear simultaneously in at least 5 images.

4. The identified specific objects or attributes represent some important contents of an individual image but not in the common objects or attributes found in the previous step.

5. Your description of each image should be simple and only 3 words.

6. Your response should be in the format of a JSON file, of which each key is a simple image index and each value is an object or attribute.

\end{ttfamily}

To enhance the quality of disentanglement and description of atoms, this task is structured into three steps, drawing inspiration from the CoT method.

\begin{ttfamily}

Step 1, take an overview of all 15 images and summarize all possible common objects or attributes that appear simultaneously in at least any 5 of these images. 

Step 2, for each individual image, identify the common objects or attributes found in Step 1 that also appear in the current image to describe the current image.

Step 3, for each individual image, you can also use some specific attributes or objects that are not common across these images to describe the current image if there is not enough 3-word description for the common object or attribute found in Step 2.

\end{ttfamily}

\subsection{Prompt for CoE Local Explanation}

In this paper, we design a prompt \texttt{prompt-coe} for the LLM to aggregate all information along the concept circuit and generate a local explanation chain to explain the decision-making process of a DVM.
This explanation chain is similar to CoT in terms of the structure of the output.
The prompt template is presented in Table \ref{tab.prompt-coe}.
The information provided in this prompt includes the DVM's prediction, sample label, image caption, relevant concept explanations derived by applying the CPDF mechanism on the automatically constructed ACD-$\mathcal{B}_{\mathcal{M}, \mathcal{T}}$ dataset, and their corresponding relevance values. 
The concept explanations and relevance values are presented in a structured format.

\begin{table}[t]
\caption{The prompt template of \texttt{prompt-coe}. 
Before querying the LLMs, we substitute the curly brackets with actual texts.}
\label{tab.prompt-coe}%
\centering
   \includegraphics[width=0.91\columnwidth]{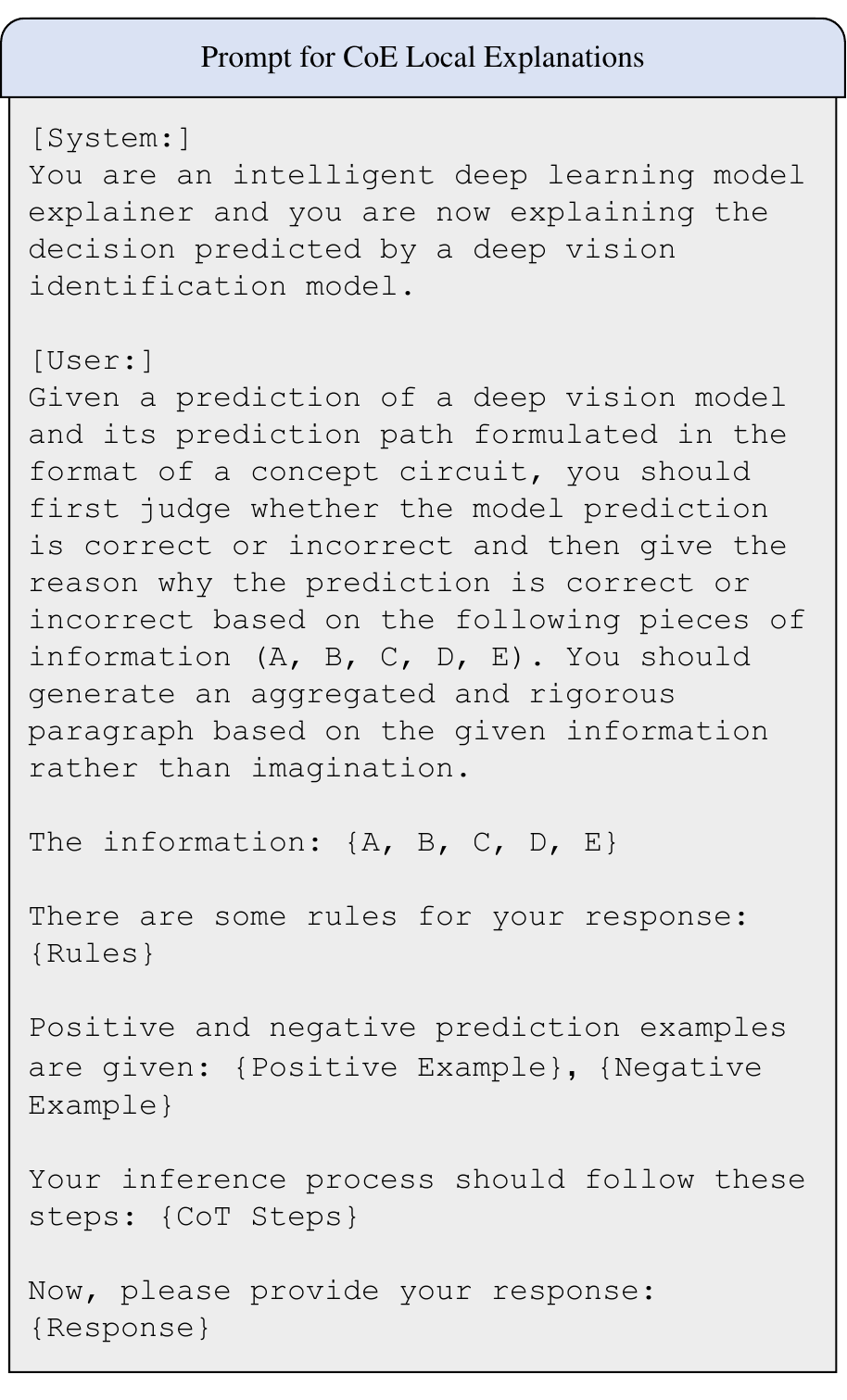} 
\vskip -0.1in
\end{table}

In this prompt, the rule set primarily functions to regulate the output. 
Furthermore, we develop two examples of local explanations based on few-shot prompting: a positive example, wherein the CoE generates local explanations corresponding to a correct prediction of the DVM, and a negative example, illustrating the expected local explanation when the DVM prediction is incorrect. 
Likewise, the process of CoE generating local explanations adheres to the CoT method, as detailed below.

\begin{ttfamily}
Step 1, Based on information A), which is the model's prediction, and information B, which is the ground truth label of the input image,  
You first need to determine whether the two are semantically equivalent.  
If they are semantically equivalent, then the model's prediction is considered correct.  
If the prediction and the label are not semantically equivalent, it is considered an incorrect prediction. 

Step 2, Based on the judgment in Step 1 and the given information C, D, and E, which include the caption of the input image,  
the vision model's decision path and the concept information at each node along the path, and the concept relevance values at each node,  
you need to explain why the model arrived at this correct or incorrect prediction.  
Analyze the decision process by examining each concept in the decision path to determine how they contributed to the final outcome. 
\end{ttfamily}

\subsection{Prompt for Evaluating Local Explanations}
\label{sec.prompt.eval}

It is essential to evaluate the generated linguistic local explanations utilizing LVLMs.
As illustrated in Table \ref{tab.prompt-coeeval}, to ensure rigor and precision, we meticulously design a prompt \texttt{prompt-coe-eval}, which primarily comprises four components: key information, evaluation criteria, evaluation steps, and rules. 
The key information includes the image, prediction, label, and the generated local explanations. The three explainability evaluation criteria—Accuracy, Completeness, and User Interpretability—are discussed in detail in Sec. \ref{sec.coeeval-local}. 
Each criterion follows a three-level scoring system (2, 1, 0).
These scoring guidelines are explicitly conveyed to the LVLM. 
The explanation process also adheres to the CoT method, requiring the LVLM to first output the scores alongside evidence and then aggregate these into a final score, as outlined below.

\begin{table}[t]
\caption{The prompt template of \texttt{prompt-coe-eval}. 
Before querying the LVLMs, curly brackets are filled with actual texts.}
\label{tab.prompt-coeeval}%
\centering
   \includegraphics[width=0.91\columnwidth]{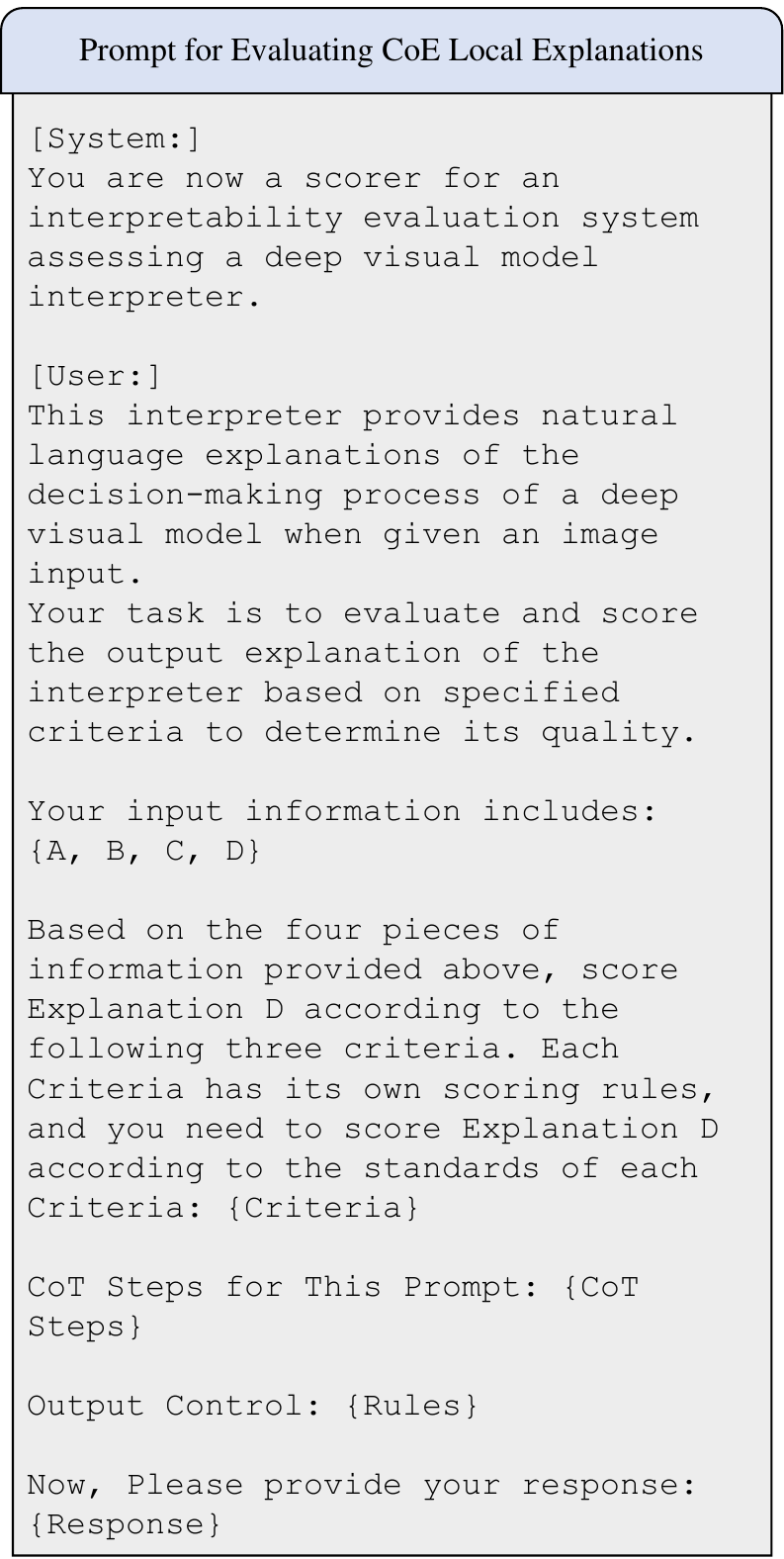} 
\vskip -0.1in
\end{table}

\begin{ttfamily}

Please first provide evidence of your evaluation for each criterion and then provide your score for each criterion, avoiding any potential bias and ensuring that the order in which the responses were presented does not affect your judgment.

Then sum up the above scores of the three criteria as the total score.

Finally, output the evidence and scores for these criteria.

\end{ttfamily}

\section{Details of the CoE Approach}
\label{sec.coeapproach}

In this section, we present supplementary details of the CoE approach, including the formulations of the CPE and the distinction from CoT.

\subsection{Probability of Concept Atoms and CPE}
\label{sec.coeother}

In this paper, we acquire the probability of concept atoms by calculating their frequency of occurrence in the disentangled concept atom set $\mathcal{A}$, given a fixed parameter $Q$ and $N$.
The naive version of probability of $i_{th}$ atom is 
\begin{equation} \label{eq.prob-naive}
    p_i^{Naive} = \frac{Num_i}{Q \times N}.
\end{equation}
Accordingly, the naive CPE of $j_{th}$ concept can be formulated as
\begin{equation} \label{eq.Hjstar-naive}
    H_j^{Naive} = \frac{-\sum_{i=1}^{Q} p_i^{Naive} \log p_i^{Naive}}{\log (Q)}.
\end{equation}
As the CPE proposed in this paper serves as an indicator of the interpretability of concepts and DVMs, we normalize the entropy value to a range between 0 and 1 by dividing it by the logarithm of the number total of atoms.

We cluster the atoms in $\mathcal{A}$, as some disentangled atoms are semantically equivalent. 
The probability and CPE of the clustered atoms in  $\mathcal{A}^\ast$ are formulated as
\begin{equation} \label{eq.prob-cluster}
    p_i^{Cluster} = \frac{Num_i}{Q \times N},
\end{equation}
\begin{equation} \label{eq.Hjstar-cluster}
    H_j^{Cluster} = \frac{-\sum_{i=1}^{P^\ast} p_i^{Cluster} \log p_i^{Cluster}}{\log (P^\ast)}.
\end{equation}

However, there exists a case in which this CPE evaluation becomes ineffective, i.e., when all image patches of a VC are highly similar, as exemplified in the first row of Table \ref{tab.appe-egmain}. 
The concept should ideally exhibit monosemanticity in this scenario. 
In contrast, the CPE calculated by Eq. \ref{eq.Hjstar-cluster} results in a value of 1, as the probabilities of all atoms are evenly distributed (\eg, each with a probability of 1/3 when $P^\ast = 3$).
To mitigate this problem, we set the minimum number of concept atoms in $\mathcal{A}^\ast$ to $N$, assuming that each VC contains at least $N$ common atoms. 
The padding atoms (in a number of $Pad = N–P^\ast$) are each assigned a frequency of 1. 
This operation preserves the relative probabilities among the $P^\ast$ atoms, ensuring that more frequent atoms remain prevalent while less frequent ones retain their lower counts.
Upon completion of these procedures, the probability and CPE are updated to Eq. 8 and Eq. 9 in the main manuscript.
The experiments are conducted in Sec. \ref{sec.sup.exp-cpe-versions}, showing the effectiveness of our method.

\subsection{Explanation of the Concept}

In this paper, we define each channel or neuron of DNNs as a VC, represented by a set of masked image patches \cite{appe-lrpsurvey, appe-clipx}.
Notably, some works consider each image patch as a VC \cite{appe-explainanything}, resembling a form of pixel-level semantic segmentation.
Channel-based interpretation can serve as both global and local explanations for a DVM by decoding and describing the commonalities among a set of image patches. 
It better represents the decision concepts learned internally by the DVM.
In contrast, the latter, identifying the key regions within a given image, only serves as a local explanation.
Our CoE approach can automatically describe these two directions, as both take the form of image patches.

\subsection{Discussion Between CoE and CoT}

The CoE approach proposed in this paper draws inspiration from CoT \cite{appe-cotsurvey1,appe-cotsurvey3}, yet with notable distinctions.
CoT directly guides large-scale models to articulate their decision-making processes through carefully crafted prompts.
However, it has the following limitations: 
1. Each step in the CoT still relies on the large model’s own capabilities, and each prediction of the current step remains unexplained; 
2. CoT mainly emerges in LLMs or LVLMs, whereas the capability for smaller DVMs is insufficient. 
In contrast, CoE dissects the DVMs by identifying critical decision concepts within key layers. 
These concepts, described in natural language, serve as nodes in a chain. 
CoE aggregates these nodes to form a coherent explanation chain that elucidates the DVM’s decision-making process. 
Although the output structure resembles that of CoT, the construction of the CoE explanation chain is achieved by leveraging the general capabilities of LVLMs to automatically describe the VCs and construct the explanation chains.
Additionally, CoE provides global conceptual explanations for DVMs while also possessing the capability to quantify polysemanticity.
Thus, CoE and CoT are notably distinct.

\subsection{Time and Cost of CoE}

CoE primarily consists of ACD, CPE, CPDF, and local explanation steps. 
ACD and CPE can be performed offline and obtained through a one-off computation process.
Building the global ACD-$\mathcal{B}$ database on ImageNet-val takes 9 hours and costs \$70. 
After that, online inference for the local explanation of a single image requires 20 seconds and costs \$0.01.
Compared to manual labor, this cost is considered acceptable.

\section{Experiments on CPE}
\label{sec.sup.exp-cpe}

In this section, we provide additional experiments across various versions of CPE, XAI methods, model architectures, and illustrative examples of CPE.

\begin{figure}[t]
\centering
    \begin{subfigure}{0.495\columnwidth}
    \centering
    \includegraphics[width=1\columnwidth]{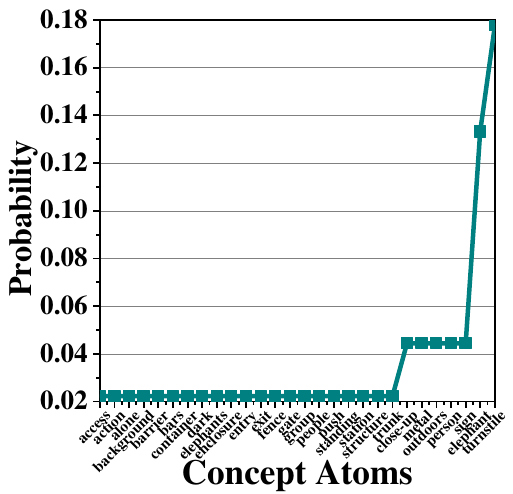}
    \caption{Naive Probability (CPE=0.91)}
    \label{fig.egnaiveclustercpe.naive}
  \end{subfigure}
  \begin{subfigure}{0.495\columnwidth}
    \centering
    \includegraphics[width=1\columnwidth]{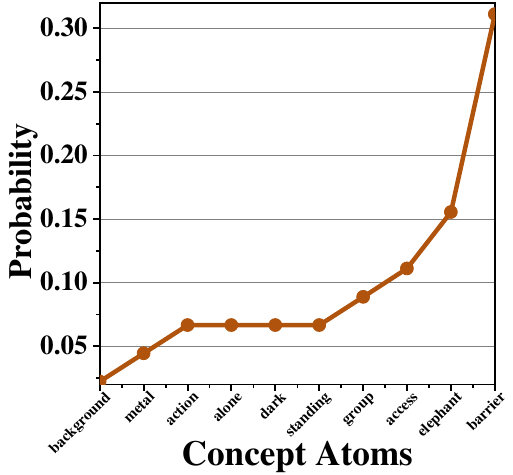}
    \caption{Clustered Probability(CPE=0.89)}
    \label{fig.egnaiveclustercpe.cluster}
  \end{subfigure}
\caption{Examples of two versions of the disentangled concept atom probability distributions. (a) shows the naive version, while (b) represents the clustered one. The channel showed here is number 163 of the output layer of Stage 4 of a ResNet152 model.}
\vskip -0.1in
\label{fig.egnaiveclustercpe}
\end{figure}

\subsection{Various Versions of CPE}
\label{sec.sup.exp-cpe-versions}

\begin{figure}[t]
\centering
    \includegraphics[width=0.8\columnwidth]{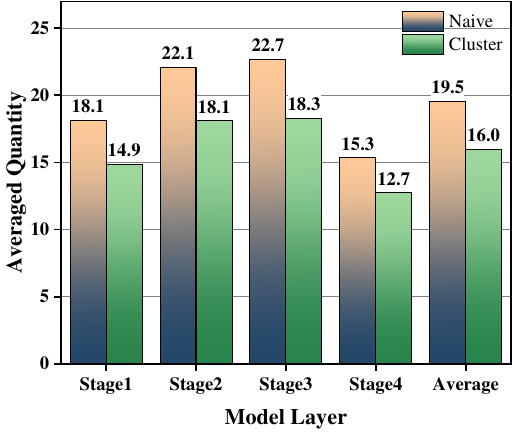}
\caption{Averaged quantity of disentangled concept atoms. The average term is the averaged quantity of 4 stages.}
\label{fig.clusteredquantity}
\end{figure}

The proposed CPE has evolved through three iterations: the naive version (Eq. \ref{eq.prob-naive} and Eq. \ref{eq.Hjstar-naive}), the clustered version(Eq. \ref{eq.prob-cluster} and Eq. \ref{eq.Hjstar-cluster}), and the final refined version (Eq. 8 and Eq. 9 in the main manuscript).
As shown in Fig. \ref{fig.egnaiveclustercpe}, some atoms disentangled for a single concept are semantically equivalent (\eg, barrier and fence, entry and gate), and many of them exhibit low probabilities. 
After clustering through the entailment model, all semantically redundant atoms are consolidated, leading to adjusted atom probabilities and a reduced CPE value.
The semantics of the atoms are mutually exclusive.
For a ResNet152 model, 3.5 atoms per concept, on average, are reduced, as illustrated in Fig. \ref{fig.clusteredquantity}.
Notably, in the third stage, where polysemanticity is most pronounced, the largest reduction is observed, averaging 4.4 per concept.
This highlights the significance of introducing the entailment model within the CPDF mechanism to cluster redundant semantics.

Furthermore, as shown in Fig. \ref{fig.clusteredquantity}, it is evident that concepts of DVMs exhibit polysemanticity, with the fewest distinct semantics occurring at the final layer (an average of 12.7 non-overlapping semantic atoms) and the most pronounced at stage 3 (18.3). 
This significantly impairs the interpretability of concepts and DVMs, and the explanations produced by concept-based XAI methods, underscoring the importance of quantifying concept polysemanticity and mitigating its impact on explanations.

\begin{figure}[t]
\centering
    \includegraphics[width=0.8\columnwidth]{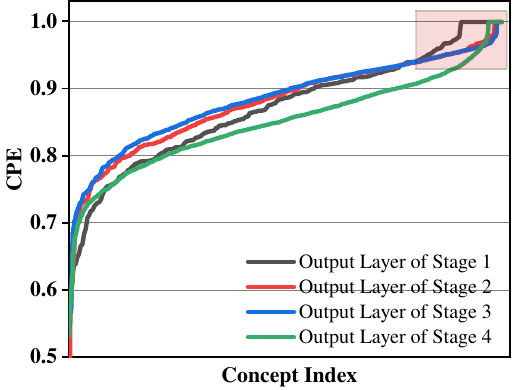}
\caption{Clustered CPEs along the channel dimension. 
The channel indices are max-normalized.
The magenta-highlighted regions emphasize results where commonalities are relatively limited, yet the computed CPE value is equal to 1.}
\label{fig.cpeonchanneldim}
\end{figure}

As shown in Fig. \ref{fig.cpeonchanneldim} and exemplified in the first row of Table \ref{tab.appe-egmain}, compared with Fig. 3(a) in the main manuscript, there exist some concepts that require padding.
Their commonalities display significant uniformity and an evenly distributed probability pattern, as discussed in Sec. \ref{sec.coeother}.
This phenomenon, where the actual level of polysemanticity is relatively low but is still calculated as high CPE, occurs more frequently in stages 1 and 4 of the DVM. 
As previously mentioned (discussed in Sec. 4.3 in the main manuscript), these stages are indeed characterized by generally lower levels of polysemanticity in their common concepts. 
After applying the padding operation, as illustrated in Fig. 3(a) in the main manuscript and the first row of Table \ref{tab.appe-egmain}, the corresponding CPE values drop to relatively low levels. 
This outcome further substantiates the effectiveness of the CPE proposed in this paper.

\begin{figure}[t]
\centering
    \includegraphics[width=0.8\columnwidth]{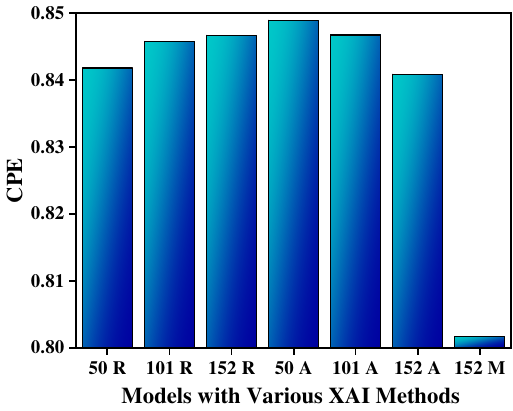}
\caption{Averaged CPE scores on different XAI methods and different models. The number under the X-axis represents the model depth. R, A, and M stand for Relevance, Activation, and Maximum mutual information-based XAI Methods, respectively.}
\label{fig.egcpexaimethods}
\end{figure}

\subsection{Different XAI Methods}

In this subsection, we conduct experiments on different XAI methods, including the relevance-based \cite{appe-naturecrp}, activation-based \cite{appe-cam}, and maximal mutual information-based methods \cite{appe-milan}. 
As illustrated in Fig. \ref{fig.egcpexaimethods}, the results reveal distinct trends in CPE across different XAI methods. 
Specifically, the polysemanticity observed in the activation-based method diminishes as model complexity increases, contrasting with the trend exhibited by the relevance-based method.
This result indicates that the choice of the XAI method $E_x$ is critical for concept-based explanations. 
Given that relevance-based concepts achieve superior fidelity and reliability in explanations compared to activation-based concepts  \cite{appe-naturecrp}, this paper adopts the relevance-based CRP as the primary $E_x$ method.
Moreover, concepts derived from maximal mutual information exhibit lower CPE values. 
However, due to the reliance on manually annotated concepts, this method lacks automation and flexibility, limiting its development on DVMs and hindering a more comprehensive comparison with other XAI methods. 
These results highlight the versatility of our approach in being applicable to various XAI methods and underscore the necessity of automating concept construction.

\begin{figure}[t]
\centering
\includegraphics[width=0.8\columnwidth]{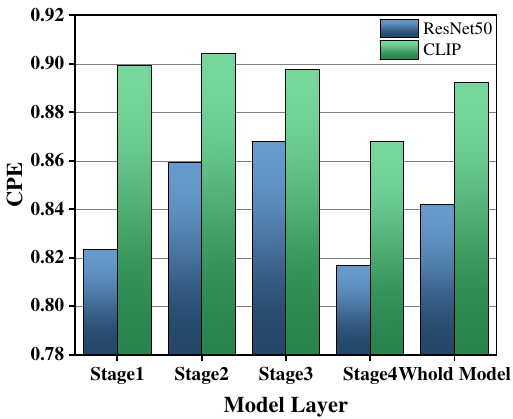}
\caption{Averaged CPE scores on 2 model architectures, i.e., ResNet50 and CLIP.}
\label{fig.cliprn50}
\end{figure}

\subsection{Different Model Architectures}

We calculate CPE values across different model architectures, including ResNet50 trained on ImageNet \cite{appe-resnet, appe-imagenet} and CLIP-ResNet50 trained on a large-scale vision-language dataset \cite{appe-openclip, appe-clip, appe-laion}. 
As presented in Fig. \ref{fig.cliprn50}, the vision branch of the CLIP model exhibits greater polysemanticity. 
The general trend aligns with that of the original ResNet, where polysemanticity is lowest in the abstract stage 4 and peaks in the intermediate stages. 
Polysemanticity is high in the shallowest stage.
We infer that these results of CLIP-ResNet50 stem from the constructed global explanation dataset ACD-$\mathcal{B}_{\mathcal{M}, \mathcal{T}}$, which is derived from the Out-of-Distributed (OOD) ImageNet Validation dataset $\mathcal{T}$ rather than the independent and identically distributed (iid) vision-language dataset utilized for CLIP's training. 
Since CLIP operates in a zero-shot mode, the representation of each VC through 15 image patches does not fully align with the conceptual requirements of the original CLIP model, resulting in increased polysemanticity.
Moreover, the vision-language dataset utilized for CLIP training encompasses significantly more categories and samples compared to the ImageNet dataset, resulting in an increase in the semantic scope that each concept must represent. 
This, in turn, amplifies the model's polysemanticity.
Our CPE method not only captures and precisely represents these phenomena through the lens of polysemanticity but also proves its effectiveness under zero-shot conditions, reinforcing the validity of the proposed approach.

\begin{table*}[t]
\caption{Additional examples of VCs and their disentangled concept atoms, along with their concept probability distributions and CPE scores. 
From top to bottom, the concepts displayed in the table are derived from concepts 141, 366, 1681, and 121 in the fourth stage of a ResNet152 model.
The CPE values increase progressively from the top row to the bottom row.}
\label{tab.appe-egmain}%
\centering
\begin{tabular}{m{11cm}<{\centering}m{4.2cm}<{\centering}|m{0.8cm}<{\centering}}
\toprule
    Visual Concept and Disentangled Concept Atoms & Concept Distribution & CPE \\
\toprule
   \includegraphics[width=1\linewidth]{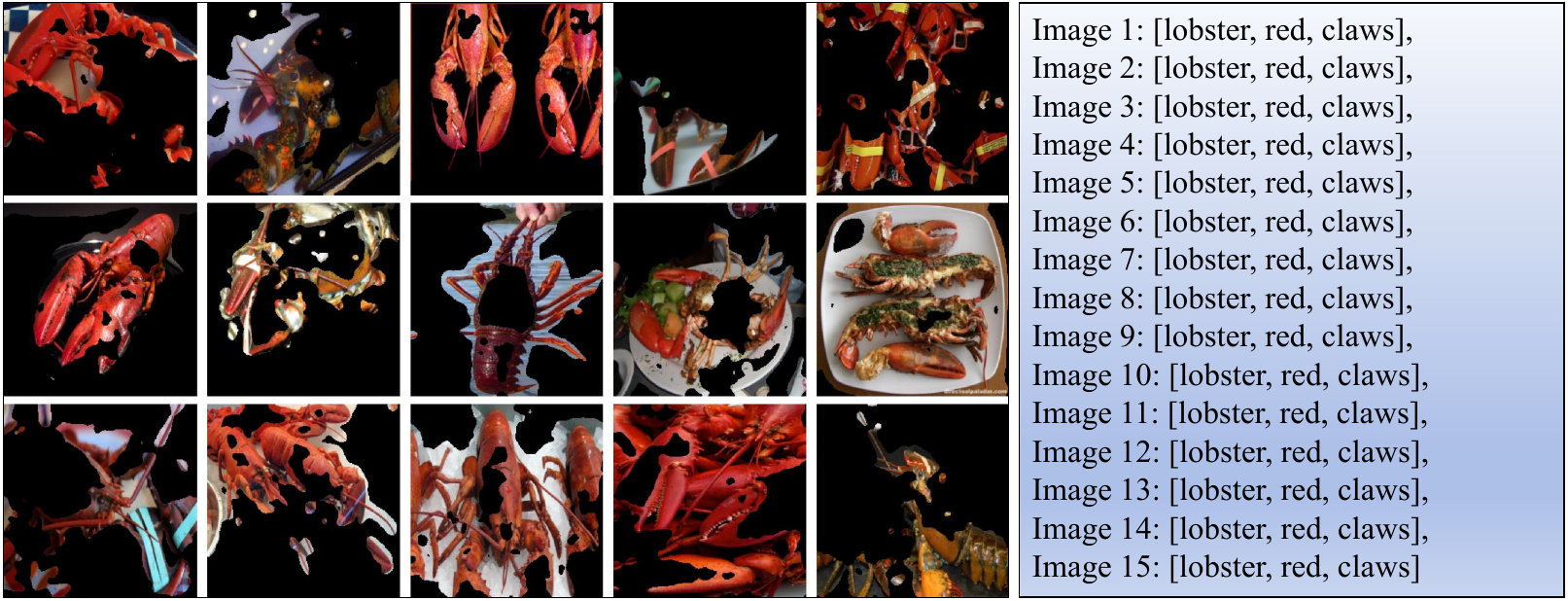} 
   & \includegraphics[width=1\linewidth]{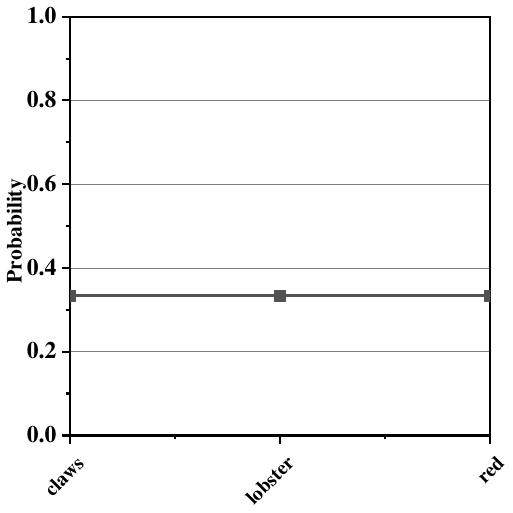}
   &\multirow{1}{*}{\textbf{0.70}}
   \\
   \midrule
\includegraphics[width=1\linewidth]{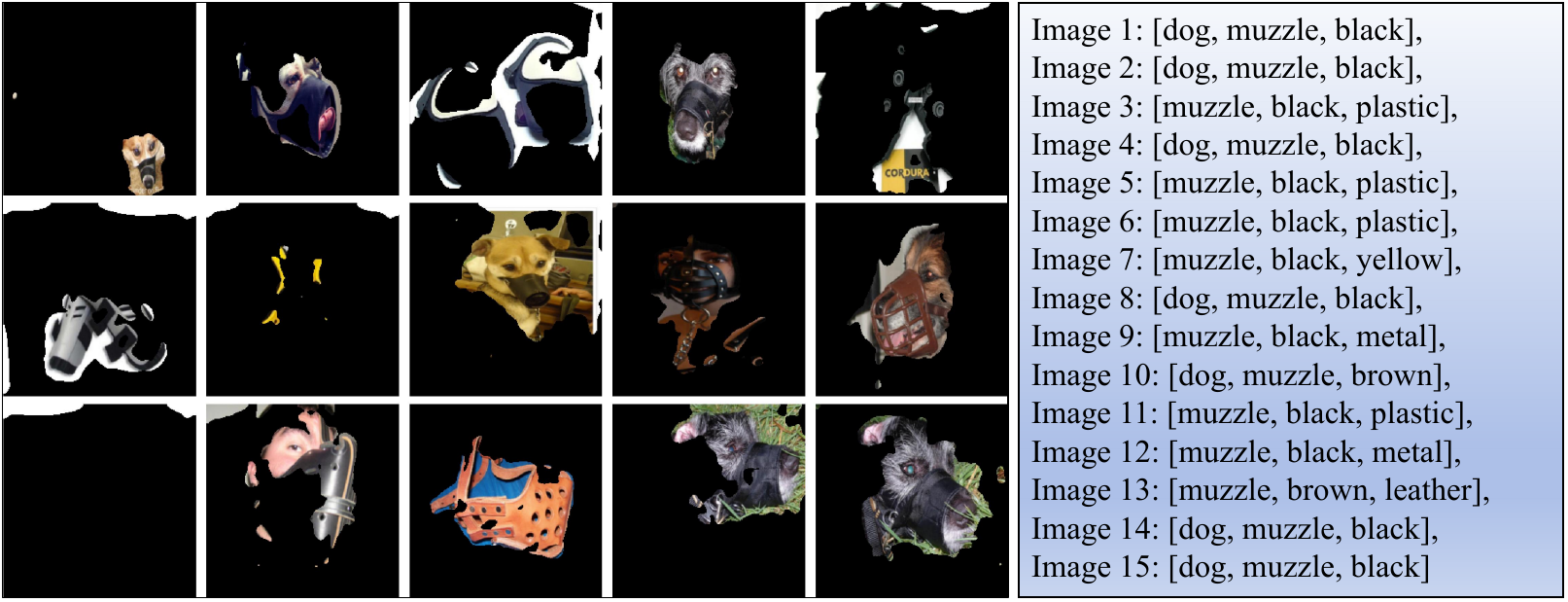} 
   & \includegraphics[width=1\linewidth]{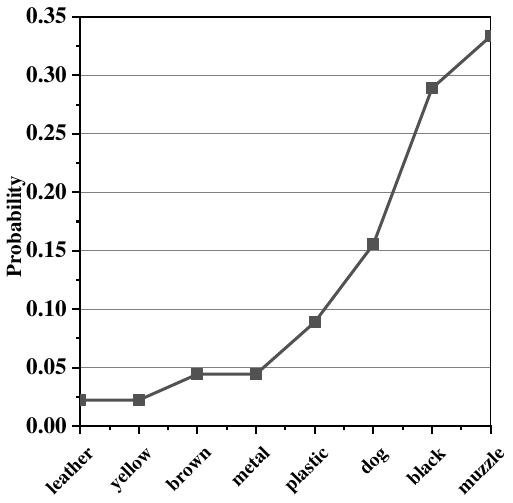}
   &\multirow{1}{*}{\textbf{0.77}}
   \\
   \midrule
   \includegraphics[width=1\linewidth]{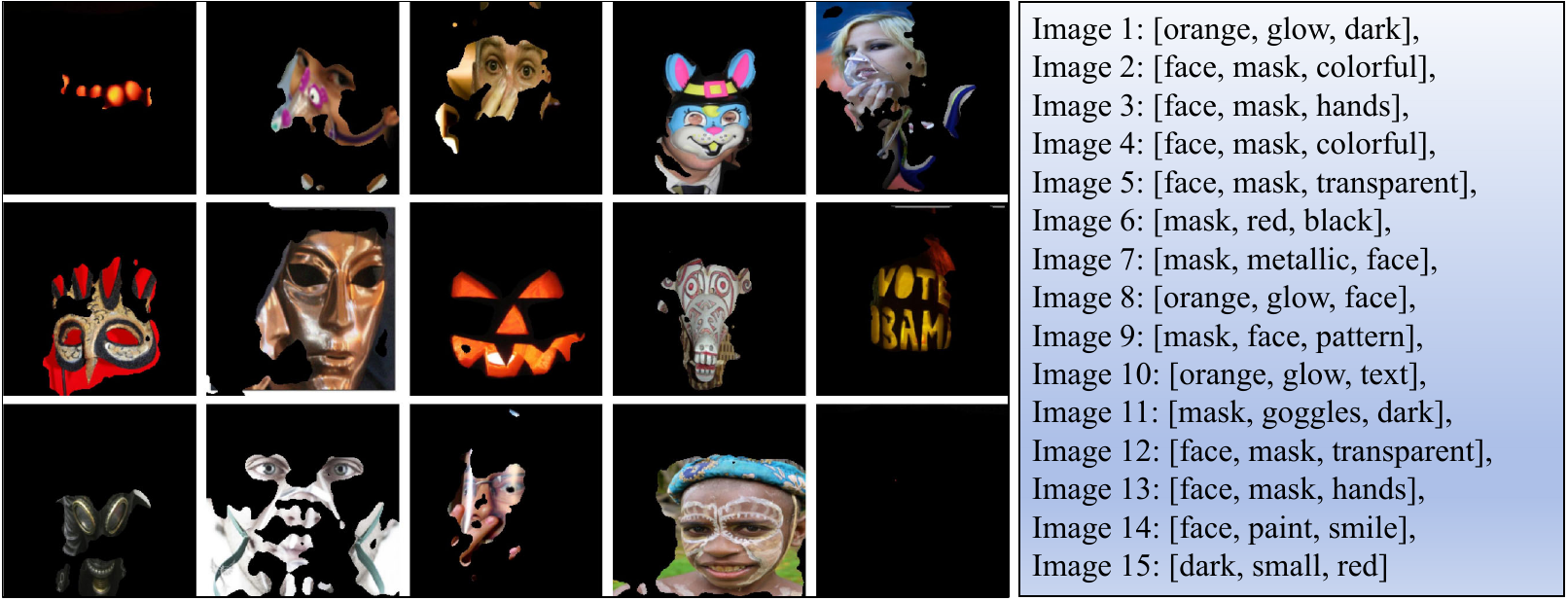} 
   & \includegraphics[width=1\linewidth]{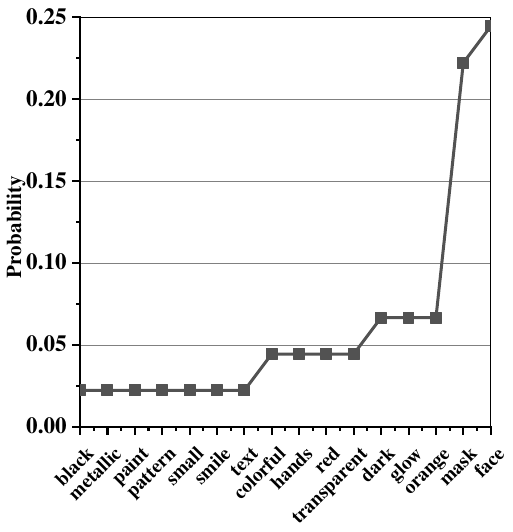}
   &\multirow{1}{*}{\textbf{0.85}}
   \\
   \midrule
   \includegraphics[width=1\linewidth]{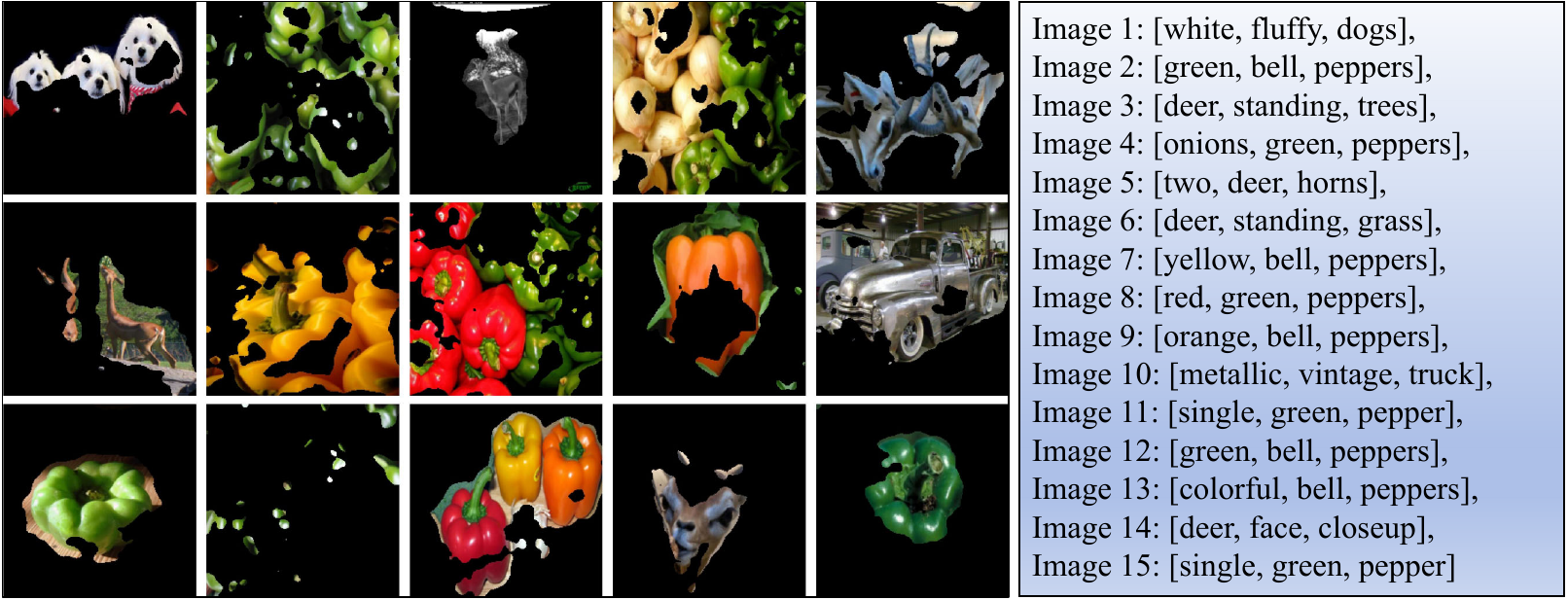} 
   & \includegraphics[width=1\linewidth]{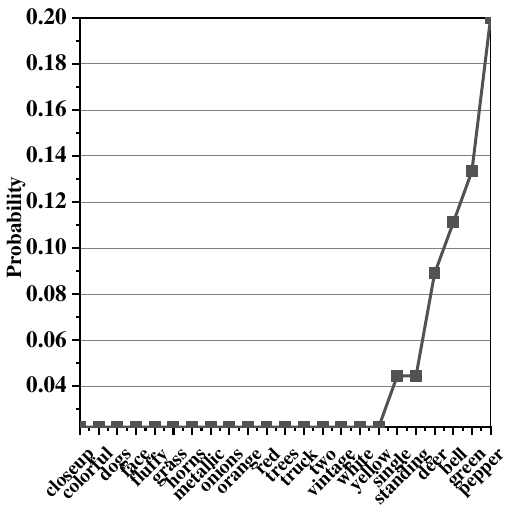}
   &\multirow{1}{*}{\textbf{0.88}}
   \\
\bottomrule
\end{tabular}%
\vspace{0.38in}
\end{table*}%

\begin{table*}[t]
\caption{Explainability metrics utilized for evaluating the local explanations. These criteria are prompted for both GPT-4o and human-based scoring systems. 
Each criterion has a maximum score of 2 points. The total explanation score is 6.
The \textbf{bolded} areas represent the core decision rationale for the scoring process.
}
\label{tab.eva_coe_local}%
\centering
\renewcommand{\arraystretch}{1.3}
\begin{tabular}{m{3.2cm}<{\centering}|m{1cm}<{\centering}|m{11.9cm}<{\raggedright\arraybackslash}}
\toprule
    Metrics & Score & Details of Each Criterion \\
\midrule
   \multirow{5}{*}{\textbf{Accuracy}}
   & \multirow{1}{*}{\textbf{2}}
   &  
\textbf{\textbf{Almost all relevant explanations}} focused on key decision points, essential features, important regions, and background information, \textbf{\textbf{with no extraneous or irrelevant content}}. \\
   \cline{2-3} 
   & \multirow{1}{*}{\textbf{1}}
   &  Explanation is \textbf{generally relevant but may contain some minor off-topic or unnecessary} information. \\
   \cline{2-3}
   & \multirow{1}{*}{\textbf{0}}
   &  Explanation includes a \textbf{significant amount of irrelevant} content, diverging from the model’s decision-making process and impairing comprehension.
   \\
\midrule
   \multirow{5}{*}{\textbf{Completeness}}
   & \multirow{1}{*}{\textbf{2}}
   &  \textbf{\textbf{Comprehensive explanation covering all major steps}}, key features, background information, and relevant concepts of the model’s decision process. \\
   \cline{2-3}
   & \multirow{1}{*}{\textbf{1}}
   &  Explanation addresses \textbf{primary decision steps but may slightly overlook some information} or secondary features. \\
   \cline{2-3}
   & \multirow{1}{*}{\textbf{0}}
   &  \textbf{Incomplete explanation lacking essential decision steps} or information, making comprehension \textbf{challenging}.
   \\
\midrule
   \multirow{5}{*}{\textbf{User Interpretability}}
   & \multirow{1}{*}{\textbf{2}}
   &  Explanation \textbf{\textbf{allows users without specialized knowledge}} to understand the model’s decision logic, \textbf{\textbf{with clear, straightforward language and smooth readability}}. \\
   \cline{2-3}
   & \multirow{1}{*}{\textbf{1}}
   &  Explanation is \textbf{mostly understandable to users with a technical background}; it is fairly clear \textbf{but may require some re-reading due to less} fluent phrasing or logic. \\
   \cline{2-3}
   & \multirow{1}{*}{\textbf{0}}
   &  Explanation is \textbf{difficult to comprehend}, with \textbf{disorganized or unclear language} that obscures the decision process of the model.
   \\
\bottomrule
\end{tabular}%
\end{table*}%

\begin{figure}[t]
\centering
\includegraphics[width=0.99\columnwidth]{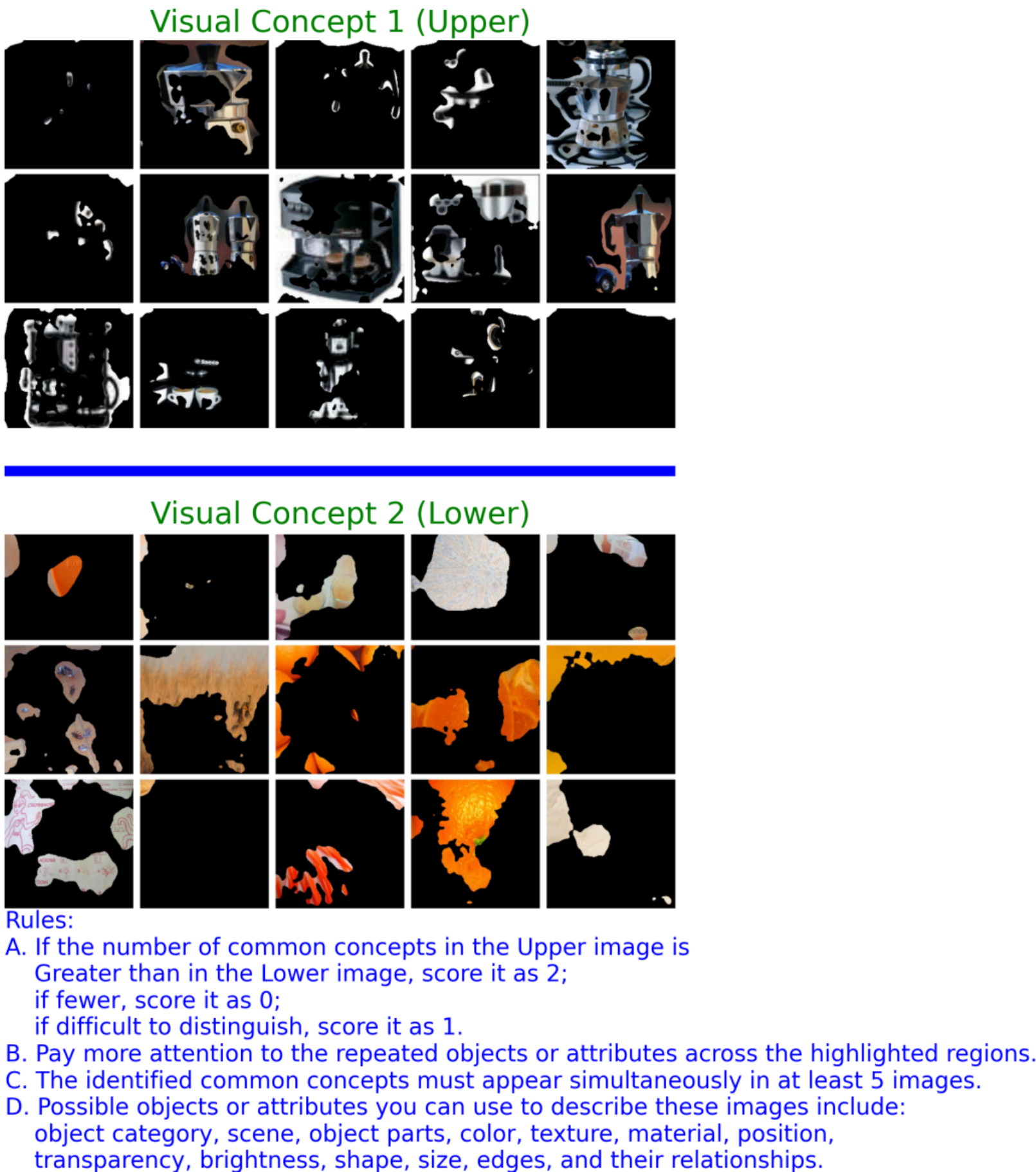}
\caption{Examples of polysemanticity comparisons between human evaluations and CPE metric.}
\label{fig.vccompare}
\end{figure}

\subsection{Human Evaluation on CPE}

In this subsection, we present a comprehensive overview of the experiments conducted to perform human evaluations on the CPE metric.
We first automatically generate a global concept explanation dataset ACD-$\mathcal{B}_{\mathcal{M}, \mathcal{T}}$ utilizing the CoE approach proposed in this paper.
By incorporating the manually annotated MILAN ANNOTATION dataset \cite{appe-milan}, we construct a comprehensive VC library for this evaluation, comprising a total of 7,680 VCs.
Recognizing the inherent difficulty for humans to directly quantify polysemanticity visually, we instead invite participants to compare the relative degrees of polysemanticity between two VCs.
From this VC library, 300 pairs of VCs are randomly sampled, each displaying varying degrees of polysemanticity. 
As illustrated in Fig. \ref{fig.vccompare}, each pair is presented to human evaluators. 
Participants are instructed to assign a score of 2 if the upper VC exhibits more polysemanticity than the lower one, a score of 1 if the opposite is true, and 1 if the distinction is unclear. 
The evaluation guidelines also prompt 13 feasible semantic directions and rules consistent with those prompted for the CPE metric.
All VC pairs are divided into 10 groups, with each group evaluated by three participants. 
If the average score exceeds 1, the upper VC is judged to have greater polysemanticity than the lower one; otherwise, it is assigned a score of 0. 
Consistency between these results and the CPE metric is then calculated, as summarized in Table 2 in the main manuscript.
The results reveal a 75\% agreement between human evaluations and the CPE metric, thereby demonstrating the validity of the CPE method.

\subsection{Examples of Disentanglement and CPE}

In this subsection, we present additional examples illustrating the disentanglement of VCs into concept atoms, as well as the probability distributions and CPE values of the clustered atoms.
As shown in Table \ref{tab.appe-egmain}, the experimental results align with the analyses presented in Sec. 4.2 in the main manuscript. 
The proposed CoE approach effectively and accurately disentangles VCs into linguistic concept atoms. 
Furthermore, the entailment model successfully clusters semantically equivalent atoms into mutually orthogonal groups, assigning corresponding probabilities to them.
The proposed CPE metric quantifies the polysemanticity of different VCs, with the subjective visual comparisons and the CPE results demonstrating consistent trends. 
The polysemanticity of the VCs in the table increases progressively from the first row to the last. Correspondingly, the disentangled atoms, their associated probabilities, and the CPE values exhibit consistent changes in alignment with this trend.
These results collectively validate the effectiveness of the approach proposed in this paper.

\begin{table*}[!h]
\setlength{\tabcolsep}{4.1pt}
\renewcommand{\arraystretch}{0.5} 
\captionsetup{font=small}
\caption{
    Comparisons of GPT-4o explanation scores under various scenarios. 
    $\textbf{$\dagger$}$: the results of baselines are obtained by applying ACD and local explanation steps, without CPDF.
    }
\label{tab.sup2}
\centering
\begin{tabular}{l|c|c|c|c}
\toprule
    Method  & Accuracy & Completeness & User Interpretability & Total Explanation \\
\midrule
    Places365 Dataset \cite{appe-places} (Baseline${}^{\textbf{$\dagger$}}$)    & 1.01 & 1.07 & 1.04 & 3.12 \\
    \rowcolor{c1!20}
    CoE on Places365 Dataset (Ours)     & 1.68 & 1.69 & 1.67 & 5.04 \\
\midrule
    Chest X-ray Dataset \cite{appe-chest} (Baseline${}^{\textbf{$\dagger$}}$)     & 1.55 & 1.62 & 1.56 & 4.73 \\
    \rowcolor{c1!20}
    CoE on Chest X-ray Dataset (Ours)   & 1.81 & 1.74 & 1.76 & 5.31 \\
\midrule
    ViT-B-16 \cite{appe-attnlrp} (Baseline${}^{\textbf{$\dagger$}}$)    & 1.18 & 1.14 & 1.16 & 3.48 \\
    \rowcolor{c1!20}\
    CoE on ViT-B-16 (Ours)     & 1.65 & 1.69 & 1.58 & 4.92 \\
\bottomrule
\end{tabular}            
\end{table*}

\begin{table}[!h]
\setlength{\tabcolsep}{2.5pt} 
\renewcommand{\arraystretch}{0.5}
\captionsetup{font=small}
\caption{
GPT-4o scores on other concept explanation methods. 
}
\label{tab.sup1}
\centering
\begin{tabular}{l|c|c|c|c}
\toprule
Method  & Acc. & Comp. & User I. & Total \\
\midrule
CLIP-Dissect\cite{appe-clipdessect} +Descrip. & 1.10 & 1.13 & 1.08 & 3.31 \\
\bottomrule
\end{tabular}            
\end{table}

\section{Experiments on CoE Local Explanations}
\label{sec.sup.exp-coe}

In this section, we elaborate on the evaluation employed to assess the linguistic local explanations.
We also present and compare additional instances of local explanations generated by CoE and other methods.

\subsection{Evaluation of Local Explanations}
\label{sec.coeeval-local}

The local explanations are evaluated from three explainability evaluation metrics, namely, Accuracy, Completeness, and User Interpretability \cite{appe-xaievaluation}.
We exclude the fidelity criterion, as CoE finds the key concepts of DVMs through existing concept circuit methods, inherently aligning its fidelity with these approaches.
Each metric is assigned three score levels: 2 points for optimal performance, 0 points for the lowest performance, and 1 point for a moderate score, as presented in Table \ref{tab.eva_coe_local}. 
The maximum score for each metric is 2, with a total possible score of 6.

\begin{figure}[!t]
\centering
\includegraphics[width=0.99\columnwidth]{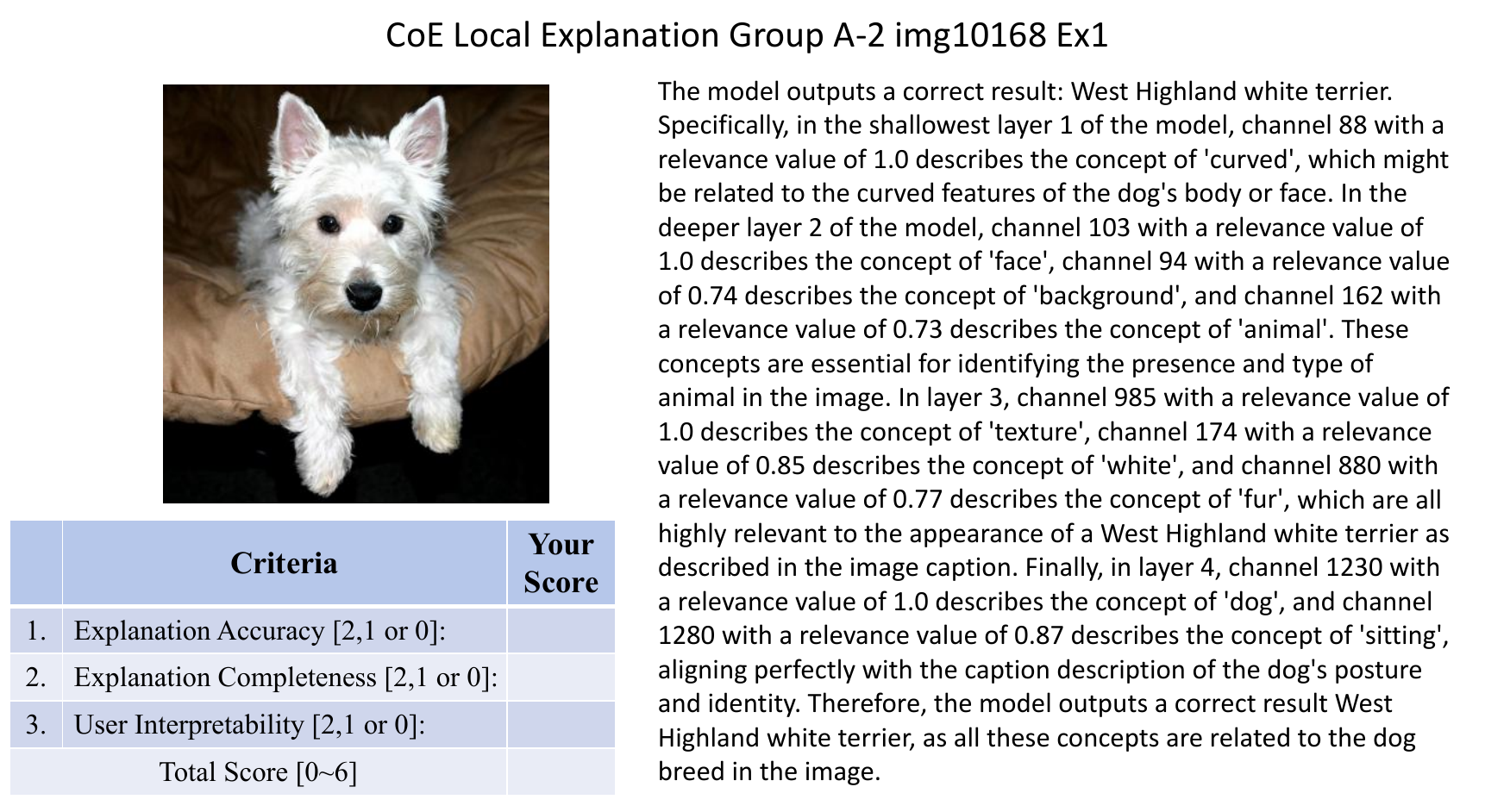}
\caption{Examples used for evaluating the local explanations generated by three methods.}
\label{fig.coeompare}
\end{figure}

These evaluation criteria are provided to both human evaluators and GPT-4o to score the generated linguistic local explanations. 
To construct the database for GPT-4o-based evaluation, 500 samples are randomly selected from the ImageNet Validation dataset. 
We sample from the correctly and incorrectly predicted instances of the DVM in a 7:3 ratio, in alignment with the accuracy rate.
Three methods, including baseline, CoE without filtering, and CoE, are evaluated in this paper.
They generate local explanations for these samples. 
The evaluation prompt for GPT-4o is discussed in Sec. \ref{sec.prompt.eval}.
Given the complexity of this evaluation for humans, we randomly select 100 samples from the former database to construct the database for human-based evaluation.
The evaluation page, as shown in Fig. \ref{fig.coeompare}, consists of the sampled image, the generated local explanations, and the scoring criteria for the three explainability metrics. 
The three methods are anonymously labeled as Ex1, Ex2, and Ex3.
As for human evaluations, the 100 samples are divided into 10 groups, with each group consisting of 30 linguistic explanations assessed by three participants.
The results, presented in Table 4 in the main manuscript, demonstrate that the CoE approach outperforms the other two methods across all three explainability metrics, confirming the superiority of the proposed approach.

\subsection{Supplemental Quantitative Evaluation Results}
\label{sec.coeeval-quantitative}

We conduct experiments of CoE on a Transformer architecture (i.e., ViT-B-16) \cite{appe-attnlrp}.
As shown in the 6th row of Table \ref{tab.sup2},
CoE is effective for the Transformer architecture, achieving an improvement of 1.44 points compared to its baseline (the 5th row).
Besides, CoE is tested on two other real-world and critical applications (i.e., Places365 \cite{appe-places} and Chest X-ray \cite{appe-chest}).
Table \ref{tab.sup2} demonstrates CoE's consistent superiority, achieving scores of 5.04 and 5.31 with improvements of 1.92 and 0.58 over their baselines (without considering polysemanticity).
In the medical dataset Chest X-ray \cite{appe-chest}, CoE achieves an explanation score of 5.31 since the category variety in this dataset is relatively small.
All the images depict the human thoracic cavity, and the differences between categories are minimal. 
This implies that the polysemanticity of concepts learned within the network is more advantageous, enhancing the explainability.
We also compare a CLIP-Dissect \cite{appe-clipdessect} method for describing concepts, published in ICLR 2023.
As shown in Table \ref{tab.sup1}, the overall explainability score is 3.31, which is clearly lower than that of CoE (5.06).
CLIP-Dissect generates a single concept atom per channel, which severely underestimates the polysemanticity issue, resulting in insufficient explanations. 
All results demonstrate the robustness and scalability of the proposed CoE approach.

\subsection{Examples of CoE Local Explanations}

In this subsection, we present additional samples to demonstrate the effectiveness of the proposed CoE approach in explaining the decision-making process of DVMs.
As presented in Table \ref{tab.appe-coelocal}, the experimental results align with the analyses in Sec. 4.4 in the main manuscript. 
Nearly all concepts are closely associated with the input images.
It infers the logical relationship between the current concept and the entire context according to its contents and relevance values.
The final output explanations accurately articulate the decision pathways underlying the DVM's predictions.

\subsection{Comparison Between CoE and Baseline}

Additionally, we provide a comparative analysis of CoE-based local explanations with that generated by the baseline method.
As exemplified in Table \ref{tab.coelocal-3methods}, the concepts provided from the baseline method exhibit inconsistencies with the input images, highlighting that disregarding polysemanticity undermines the comprehensibility of local explanations. 
In contrast, the CoE approach generates linguistic explanations that encapsulate all indispensable relevant information for identifying a hog, such as the concept of pig and pink.
These results collectively validate the effectiveness and superiority of the CoE approach proposed in this paper.

\newpage

\begin{table*}[!t]
\caption{Additional cases of local explanations generated from the CoE approach. The first three rows show the explanations of correct predictions of the DVM, and the fourth row is reversed.
The yellow highlighted regions illustrate the logical relationships identified by the CoE approach between concepts within the explanation chains and between concepts and their contexts. 
The green highlighted regions represent the CoE approach's final summary and commentary on the explanation chains.}
\label{tab.appe-coelocal}%
\centering
\setlength{\tabcolsep}{1mm}{ 
\begin{tabular}{c}
\toprule
    \includegraphics[width=0.99\linewidth]{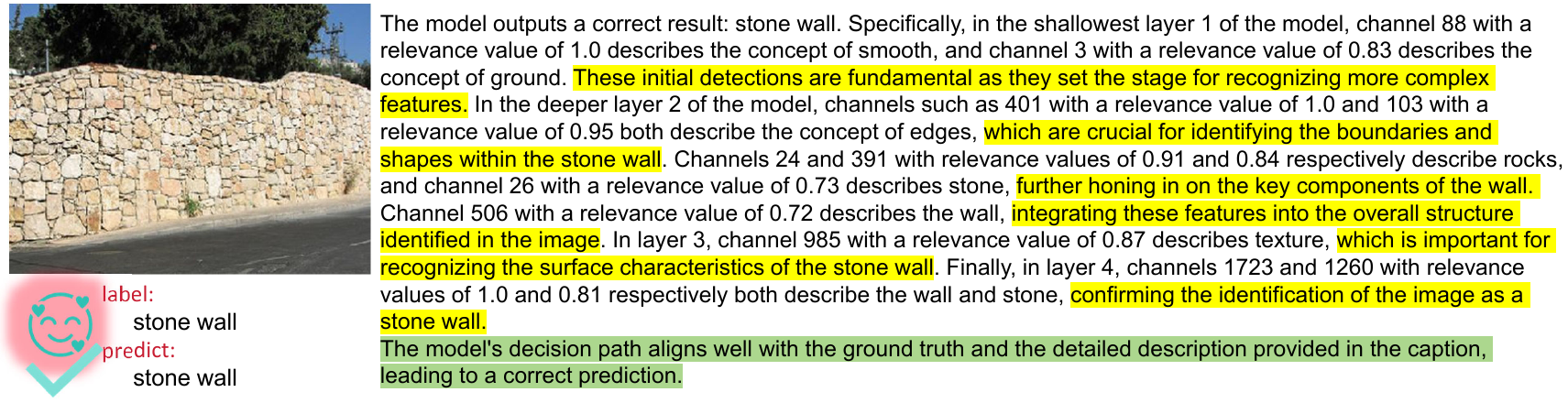}\\
    \midrule
    \includegraphics[width=0.99\linewidth]{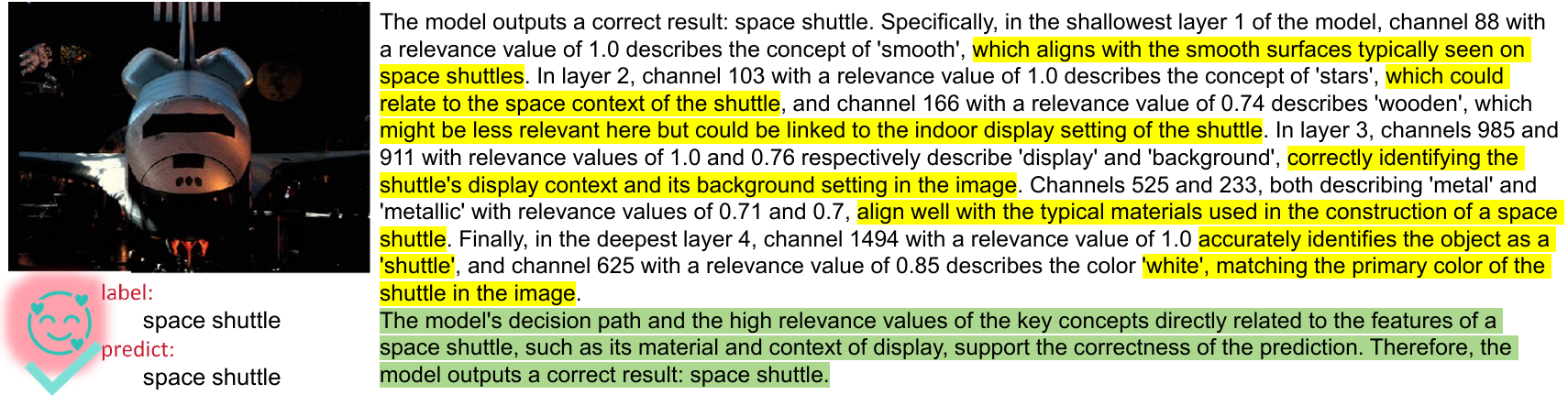}\\
        \midrule
    \includegraphics[width=0.99\linewidth]{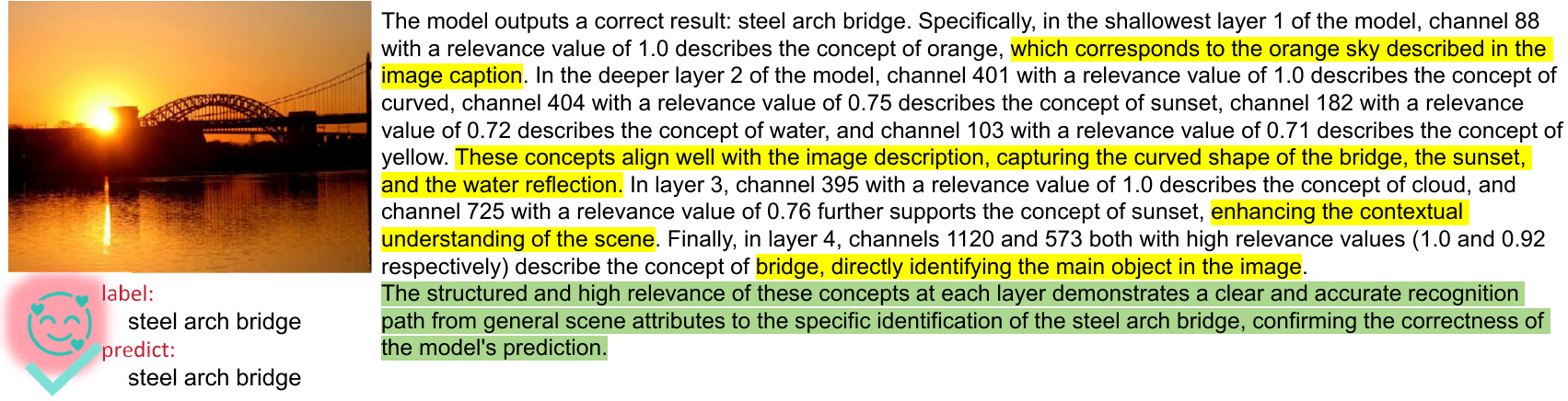}\\
        \midrule
    \includegraphics[width=0.99\linewidth]{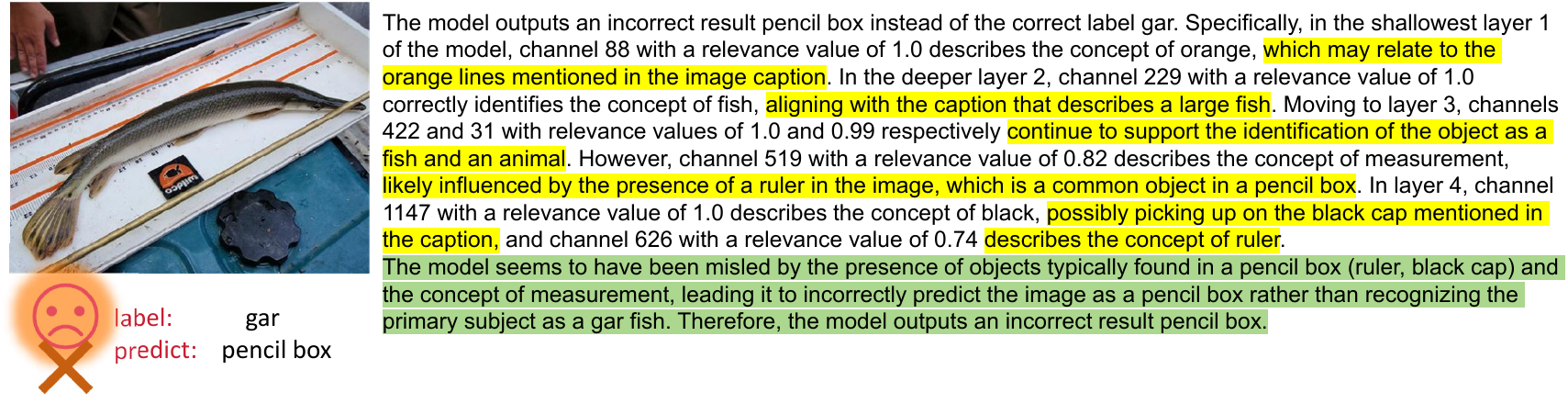}\\
\bottomrule
\end{tabular}%
}
\end{table*}%

\begin{table*}[!t]
\caption{Comparisons between local explanations generated by the baseline method and our CoE approach. 
The highlighted regions indicate the key concepts identified by different methods as influential in the model's decision-making process.
The local explanations in the first row are generated by the baseline method using the manually annotated MILAN ANNOTATION dataset, while the second row represents the outputs of our CoE approach applied to the automatically constructed ACD-$\mathcal{B}_{\mathcal{M}, \mathcal{T}}$ dataset.}
\label{tab.coelocal-3methods}%
\centering
\begin{tabular}{m{2cm}<{\centering}|m{14.2cm}<{\centering}}
\toprule
MILAN ANNO. (Baseline) &    
\includegraphics[width=0.99\linewidth]{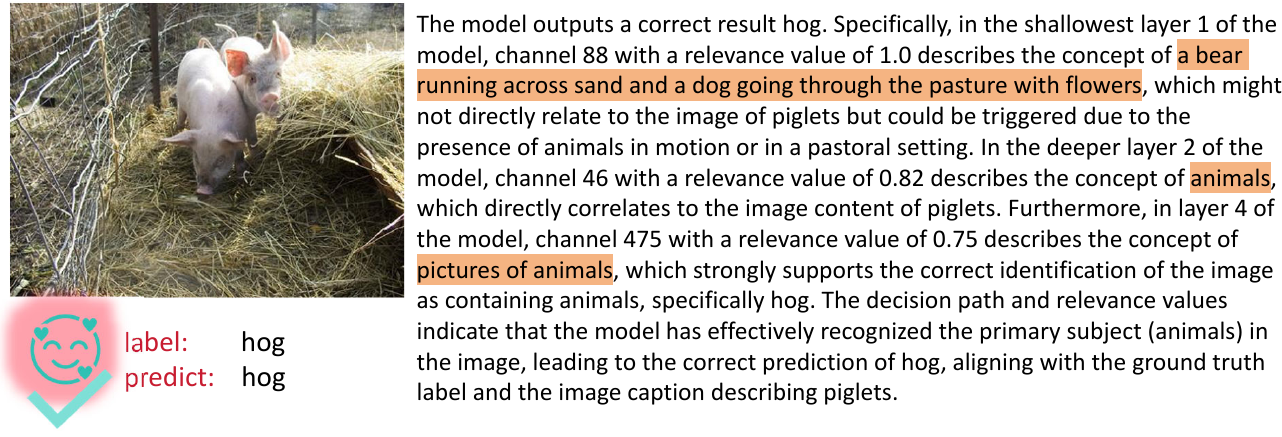}\\
    \midrule
CoE (Ours) &
    \includegraphics[width=0.99\linewidth]{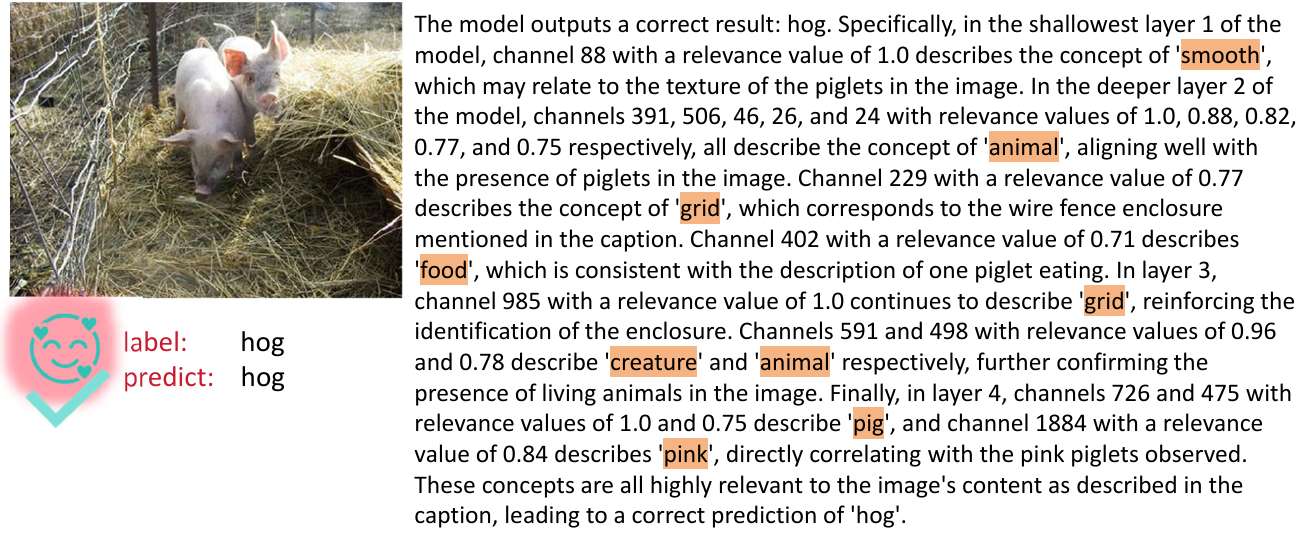}\\
\bottomrule
\end{tabular}%
\end{table*}%